\documentclass[lettersize,journal]{IEEEtran}
\usepackage{amsmath,amsfonts}
\usepackage{algorithmic}
\usepackage{algorithm}
\usepackage{array}
\usepackage[caption=false,font=normalsize,labelfont=sf,textfont=sf]{subfig}
\usepackage{textcomp}
\usepackage{stfloats}
\usepackage{url}
\usepackage{verbatim}
\usepackage{graphicx}
\usepackage{cite}
\usepackage{lineno,hyperref}
\modulolinenumbers[5]
\usepackage{algorithmic}
\usepackage[dvipsnames]{xcolor}
\usepackage{mathtools}
\usepackage{enumerate}
\usepackage{booktabs}
\usepackage{color}
\newcommand{\mat}[1]{\boldsymbol{#1}}
\newcommand{\real}{\mathbb{R}}
\newcommand{\norm}[1]{\left\lVert#1\right\rVert}

\begin{document}

\title{Connections between Deep Equilibrium and Sparse Representation Models with Application to Hyperspectral Image Denoising}

\author{Alexandros Gkillas ~\IEEEmembership{Student Member,~IEEE},
Dimitris Ampeliotis, ~\IEEEmembership{Member,~IEEE},
Kostas Berberidis, ~\IEEEmembership{Senior Member,~IEEE}

\thanks{This  work  was  supported in part by the University of Patras  and the  RPF,  Cyprus, under  the  project  INFRASTRUCTURES/1216/0017  (IRIDA).}
}



\maketitle

\begin{abstract}

In this study, the problem of computing a sparse representation of multi-dimensional visual data is considered. In general, such data e.g., hyperspectral images, color images or  video data consists of  signals that exhibit strong local dependencies.
A new computationally efficient sparse coding optimization problem is derived by employing regularization  terms that are adapted  to the properties of the signals of interest. Exploiting the merits of the learnable regularization techniques, a neural network is employed to
act as structure prior and reveal the underlying signal dependencies. To solve the optimization problem Deep unrolling and Deep equilibrium based algorithms are developed, forming highly interpretable and concise deep-learning-based architectures, that process the input dataset in a block-by-block fashion. Extensive simulation results, in the context of hyperspectral image denoising, are provided, which demonstrate that the proposed algorithms outperform significantly other sparse coding approaches and exhibit superior performance against recent state-of-the-art deep-learning-based denoising models. In a wider perspective, our work provides a unique bridge between a classic approach, that is the sparse
representation theory, and modern representation tools that are
based on deep learning modeling.

\end{abstract}


\section{Introduction}

\IEEEPARstart{O}{ver}  the past years, the sparse representation theory has evolved into a mature and highly influential mathematical modeling framework, which has led to remarkable results in a wide variety of applications across numerous disciplines e.g.,  signal processing, image processing \cite{Elad2010} and machine learning \cite{deep_elad}. A plethora of works have utilized the sparse representation framework as an effective prior to model signals encountered in various problems, ranging from image denoising \cite{Elad2006}, inpainting \cite{inpainting} and  spatial/spectral super-resolution \cite{Yang2010}, \cite{myICIP} to unmixing \cite{sunsanltv}, classification \cite{classification} and compression \cite{compression}. Building upon its elegant theoretical foundation, sparse representation modeling seeks to discover the inherent sparsity structure that exists in many natural signals \cite{Elad2010}. In greater detail, in its most usual form, this model aims to approximate a signal, represented by a vector, as a linear combination of a limited number of columns, termed atoms, from a given overcomplete matrix, known as a dictionary \cite{Tosic2011}. 

In general, sparse coding algorithms can be divided into two main categories, that is,  greedy methods and convex relaxation based approaches \cite{Bao2016}. Greedy methods try to minimize the $l_0$ pseudo-norm, that captures the sparsity of the solution, in a greedy fashion. They include prominent algorithms such as the orthogonal matching pursuit (OMP) \cite{omp}, the batch-OMP \cite{batch_omp} and the compressive sampling matching pursuit (CoSaMP) \cite{cosamp}. On the other hand, convex relaxation based approaches, such as the basis pursuit \cite{basis} and the least absolute shrinkage and selection operator (Lasso) \cite{lasso}, replace the $l_0$ pseudo-norm with the $l_1$ norm, thus forming a convex problem which is much easier to solve. 

However, with the exception of  few works \cite{myICIP}, \cite{mypaper}, \cite{sunsanltv}, most existing sparse coding algorithms do not take into account the fact that, in multi-dimensional visual data applications \cite{video_and_image2}, the signals  exhibit strong local dependencies among each other, since they treat each signal independently, limited to capture only the structure  in each
individual signal vector. One of the key points in this work is that the proposed algorithms take into account such local dependencies among signals that appear at nearby columns inside the data matrix. In the remaining of this work, we use the term data matrix to refer to a structure where its columns correspond to the two spatial dimensions (e.g., stacked) of the visual data and its rows correspond to the temporal or spectral dimension. Moreover, we use the term dependency \emph{across} the data matrix, to emphasize the fact that signals that appear at different columns inside a data matrix exhibit strong dependencies. Also, in the following,  the term  dependency \emph{along} the data matrix will  refer to any structural similarity (including sparsity) that may be present in each individual signal vector (i.e., column of the data matrix).
Signals that fall in to the category of interest in this work often appear in various engineering disciplines, e.g., image and video processing \cite{video_and_image1, video_and_image2}, remote sensing \cite{remote_sensing}, seismic data \cite{Hingee2011}.
A typical case of such signals is hyperspectral images (HSI), where each individual hyper-pixel/spectrum has internal structure (e.g., it admits a sparse representation) but also hyper-pixels located at neighbouring spatial locations demonstrate strong dependencies \cite{sunsanltv},
\cite{tsinos} (see, also Fig. \ref{fig:example_hyper}). Additionally, in numerous scenarios and settings these signals are corrupted by noise and/or interference. For instance, in remote sensing applications the quality of hyperspectral images can be degraded by several factors e.g., atmospheric turbulence, extreme temperatures and sensor imperfections \cite{16Rasti2018, Rasti_review}. Under the assumption that the considered signals exhibit dependencies both along and across the corresponding data matrix, this \emph{a-priori} information can be utilized to derive sparse coding algorithms offering enhanced signal restoration performance, but also reduced computational complexity.

To capture the local dependencies present in a data matrix, in this work, we explore the potential of an emerging body of studies which employs regularization terms properly learnt from the data via the use of suitable neural networks \cite{learned_regularizers, Zhangcvpr, dl_review_imaging}. In particular, these regularizers can be learnt effectively from a collection of training data, thus enabling them to capture the inherent structure of the data, and in the sequel, they can be properly used in an optimization problem to promote the properties of the signals of interest. Convolutional Neural Networks (CNNs) constitute effective models for learning the structure of locally dependent data and deriving proper regularization terms, due to their powerful representation capacity \cite{Zhangcvpr}, \cite{DnCNN}.

Based on these remarks and different from the above-mentioned studies, in this study, we investigate the idea of combining the learnable regularizer (i.e., a convolutional neural network) in conjunction with the sparsity promoting $l_1$ norm and a data-consistency term, to form a novel cost function for the sparse coding problem, that is able not only to reveal the sparsity nature of the signals but also to model their dependencies. Employing \emph{variable splitting} techniques, and in particular the half quadratic splitting (HQS) approach \cite{hqs}, a novel iterative sparse coding solver is formed.

Having derived an efficient solver for the problem at hand, in the sequel, we leverage our results by employing the recently developed model-based deep learning theory, and in particular, Deep Unrolling (DU) \cite{eldar} and Deep Equilibrium (DEQ) \cite{DEQ} approaches.
In particular, by unrolling a small number of iterations of the proposed solver, a deep learning architecture is formed where each layer corresponds to an iteration of the sparse coding solver. The forward pass for this network is equivalent to iterating the considered algorithm a fixed number of iterations \cite{eldar}. However, the deep unrolling techniques are characterised by several limitations, including stability, memory and numerical issues during the training process, thereby the number of the unrolling iterations must be kept quite small \cite{rebeca}, \cite{eldar}. To surmount the key limitations introduced by the DU method, a second novel sparse coding approach is proposed utilizing the efficiency of the Deep Equilibrium models. The DEQ approach aims to express the entire deep learning architecture derived from the DU method as an equilibrium (fixed-point) computation, corresponding to an efficient network with an infinite number of layers.

Adopting a wider perspective, our study aims to establish a solid and useful bridge between a classical approach, that is the sparse representation theory, and novel representation tools which are based on deep learning modeling. Similar efforts, from a different perspective,  were made in \cite{deepKSVD}, where the benefits of  connecting the problem of dictionary learning  with the more recently derived deep learning approaches were highlighted.   
In this work, we contribute to this connection considering that the proposed sparse coding algorithm can be effectively embedded into deep networks based on the deep unrolling and deep equilibrium strategies, and be trained via end-to-end supervised learning. A great benefit that stems from such an approach is the fact that the architecture of the proposed deep networks is highly interpretable, since the network parameters (i.e., the weights of the CNN prior and the regularization coefficients) derive directly from the parameters of the proposed sparse coding iterative optimization scheme. Along with their high interpretability, the derived deep unrolling and deep equilibrium models are more concise, requiring less training data. This stems  from the fact that the new models take into account the underlying physical processes and utilize prior domain knowledge in the form of correlation structure and sparsity priors.

The contributions of this work are summarized as follows:\begin{itemize}
\item Considering the potentials of the learnable regularization techniques, a novel sparse coding optimization problem is proposed, involving a sparsity promoting $l_1$ norm, a learnable regularizer (using  a convolutional neural network) and a data-consistency term. The combination of the $l_1$ norm and the CNN module empowers the proposed scheme to reveal and utilize both sparsity and structural priors adapted to the properties of the  signals, thus offering significant performance gains, especially when the signals are corrupted by severe noise.  
\item Based on the proposed sparse coding optimization framework, two novel highly interpretable deep learning-based methods are derived for solving the sparse coding problem. The first method is based on the deep unrolling paradigm, which unrolls a fixed number of iterations of the proposed  optimization scheme, thus creating a highly interpretable deep learning architecture.
The second sparse coding method utilizes the deep equilibrium models, thus expressing effectively the above deep unrolling architecture as an equilibrium computation, which corresponds to a deep network with infinite number of layers.
Unlike the deep unrolling architecture, in which the number of layers must be kept relatively small due to stability issues, the deep equilibrium model is able to provide more accurate results, overcoming efficiently many of the limitations introduced by the deep unrolling model.
Moreover, based on the findings on our previous work \cite{mypaper}, the proposed sparse coding deep networks can be significantly enhanced in terms of computational complexity, without sacrificing accuracy. In more detail, under the assumption that neighboring signals can be represented using the same support set from a given dictionary, two additional approximate sparse coding networks are derived that utilise the deep unrolling and deep equilibrium models, respectively.

\item Overall, this study provides a novel connection between a mathematically solid and extensively studied approach, i.e., sparse representation modeling, from the one side and modern deep learning methods from the other. The resulting techniques consist of highly interpretable and concise deep networks, as they are derived by utilizing prior domain knowledge for the examined problems. 
Apart from the concise nature of the proposed sparse representation networks, these approaches exhibit better hyperspectral denoising performance against several state-of-the-art deep learning architectures, providing valuable evidence that the sparse representation models in the form of deep networks can play again a central role in image and signal processing fields.
\end{itemize}

The remainder of this paper is organized as follows. In Section \ref{sec:related}, a detailed literature review of related works is given. In the sequel, Section \ref{problem_formulation} formulates the problem under study. Sections \ref{methodA} and \ref{method B} derive the proposed full and approximate algorithms, respectively. Section \ref{Results} presents a series of extensive numerical results in the context of hyperspectral image denoising, that demonstrate the efficacy of the new algorithms. Finally, Section \ref{conclusions} concludes the paper.

\section{PRELIMINARIES AND RELATED WORKS} \label{sec:related}

\subsection{Sparse coding algorithms} \label{reviewSparse}
 When signals that exhibit strong local dependencies are to be modelled (i.e., signals that demonstrate dependencies both across and along their data matrix), existing sparse coding algorithms such as the OMP, Lasso or their accelerated learnable versions (i.e., the learnable OMP \cite{lomp} and the learnable LISTA \cite{lista}) are incapable of capturing the dependencies along the data matrix, since they treat each signal vector independently. In literature, only few studies have explored the design of sparse coding algorithms by incorporating priors regarding also the dependencies of the signals along the data matrix. Among them, the most prominent algorithm, called SUnSAL-TV \cite{sunsanltv} combined the $l_1$ norm with a total variation regularizer \cite{Chambolle2004}, achieving state of the art results in the hyperspectral unmixing problem. However, this algorithm is excessively expensive in computational terms  inducing its limited applicability in real-time and/or high-dimensional applications. 
 
 Additionally, in our previous work \cite{mypaper}, a block-wise sparse coding strategy was employed under the assumption that a block of nearby signals (e.g., a small spatial patch in a hyperspectral image) can be described by the same atoms from a given dictionary (same support set). Having identified the proper support set for the signals, a total variation regularized optimization problem was proposed to compute the optimal representation coefficients. However, the total variation regularizer has a tendency to \textit{over-smooth} the reconstructed signals.

In contrast to the above approaches, in this study we argue that more accurate and computationally efficient sparse coding methods can be derived by employing regularization terms that are adapted to the properties of the signals of interest. Exploiting the merits of the learnable regularization techniques, a convolutional neural network  is employed to act as structure prior and reveal the underlying
signal local dependencies, thus forming a novel sparse coding problem. This optimization problem can be properly handled in the context of deep unrolling and deep equilibrium approaches.

\subsection{Deep Unrolling models}
The literature regarding the deep unrolling paradigm is rich, including numerous studies that aim to solve various problems, such as \cite{du1, du3, du6, du5, du4, du7, du8, du9}. Especially, in applications characterized by limited available data (e.g., MRI reconstruction), these approaches achieve state-of-the-art results \cite{mri_sota}. More formally,  the deep unrolling models convert classical iterative inverse imaging solvers into meaningful and highly interpretable deep learning architectures, where each iteration of the solver corresponds to one layer of the network. 

Our study aligns with the above works only with this unfolding strategy, since here we focus on designing sparse representation frameworks able to capture the sparsity and the inherent dependencies of signals that exhibit a  specific structure (e.g., hyperspectral images that consists of locally dependent patches).    
Note that, as we mentioned in Section (\ref{reviewSparse}) the learnable versions of the OMP \cite{lomp} and lasso \cite{lista}, although they follow the deep unrolling paradigm, however they focus only on accelerating the sparse coding procedure, without offering any additional performance accuracy. In our study, we explore interpretable deep architectures that both improve the accuracy and the computational complexity of the sparse coding process. 

\subsection{Deep Equilibrium Models}

Deep equilibrium  (DEQ) models have recently  appeared in literature, proposing an appealing framework for employing  infinite-depth networks by expressing the entire deep architecture as an equilibrium computation  \cite{DEQ}. In this study, we explore the potential of the above idea to alleviate some of the drawbacks
associated with the deep unrolling models.
To better explain the deep equilibrium approaches, let us proceed with a short introduction.
Consider a generic K-layer deep feedforward model  expressed by the following recursion
\begin{equation}
    g^{(k+1)}=f^{(k)}_\theta(g^{(k)}, y) \,\, \, for\,\, \, k=0,1 \ldots K-1,
\end{equation}
where $k$ is the layer index, $g^{(k)}$ denotes the output of the $k$-th layer, y is the input, $f^{(k)}_\theta$ stands for some nonlinear transformation. 
Interestingly, recent studies that impose the same transformation $f_\theta$ in each layer, namely $f^{(k)}_\theta = f_\theta$ were able to yield competitive results against other state-of-the-art methods \cite{wt1, wt2, wt3}. Under this weight tying practice \cite{w8tying1}, the authors in \cite{DEQ} proposed a DEQ model aiming to efficiently find the fixed point $g^\star$ where the further application of the nonlinear transformation or iteration map $f_\theta$ does not alter its value. Particularly,  the fixed point derives from the solution of the following system\begin{equation}
    g^{\star}=f_\theta(g^{\star}, y).
\end{equation}
Note that the above solution can be interpreted as an infinite depth network. However, instead of computing this point by repeating the transformation $f_\theta$, the DEQ method employs more efficient ways to obtain the equilibrium or fixed point, such as root finding techniques. 
Importantly, the weights $\theta$ of the network can be obtained via implicit differentiation using only constant memory. 

In literature, only the study \cite{rebeca} has explored the applicability of this methodology to solve generic inverse problems concerning the image reconstruction, such as MRI reconstruction or image deblurring. Different from this approach, we extend the potential of the deep equilibrium framework by providing a novel bridge between the sparse coding problem and this new deep learning method, that is the DEQ models. In more detail, the proposed sparse coding problem can be considered as a fixed-point network that is able to surpass all the limitations introduced by the corresponding deep unrolling sparse coding model. \textit{To the best of the authors' knowledge, this is the first study that investigates a connection between the sparse representation theory and deep equilibrium models.}

\subsection{Hyperspectral denoising}

Over the years, great efforts have been devoted to the problem of HSI denoising. The literature on this problem is rich ranging from studies that employ optimization-based methods, such as full-rank and low-rank approaches, to studies that employ very deep learning architectures \cite{Rasti_review}. Concerning the optimization-based  full-rank approaches, this category of studies aims to capture the spatial and spectral dependencies of the hyperspectral images by employing wavelet-based methods e.g., \cite{wavelet, fordn},  and spatial and/or spectral handcrafted regularizers such as \cite{t1,t2, octv}. The low rank approaches, in turn, have proven to be very efficient in utilizing the high spectral dimension of the hyperspectral data\cite{Rasti_review}, thus employing low-rank constraints (e.g., the nuclear norm) or a combination of low-rank constraints with other regularizers, such as the total variation and the  $l_1$ norm \cite{TDL, itsreg, LLRT, LRMR, LRTV, NMoG, TDTV, fasthyde, hyres, 8737679, 8760540}.

Recently,  leveraging the advances in deep learning, several deep architectures have been proposed \cite{QRU3D, HSID-CNN, 8693549}. The 3D Quasi-Recurrent Neural Network (QRNN3D) \cite{QRU3D} employs 3D convolution components and quasi-recurrent pooling functions to capture the spatio-spectral dependencies of the HSI, providing state-of-the-art restoration results. Also, in \cite{HSID-CNN} a spatio–spectral deep residual CNN was proposed, utilizing 3D and 2D convolutional filters to capture the dependencies of the images. 
Furthermore, as pointed out in \cite{QRU3D, Hyde}, the  MemNet \cite{MemNet} network  and one variation of it, called MenNetRes \cite{Hyde} (i.e., a combination of MemNet with the Hyres approach \cite{hyres} ) are able to provide competitive hyperspectral denoising results.

As already mentioned, the literature on hyperspectral denoising can be categorized into two major directions i.e.,  the conventional (optimization-based) and the deep-learning-based approaches. In this work, adopting a new approach, we explore a bridge between these two categories. In particular, we depart from the ad-hoc, intuition based design of deep learning models, as those mentioned above,  and move towards well-justified architectures derived from the  prior domain knowledge of the examined problem.
By providing a novel connection between the classical sparse coding problem and the deep equilibrium and deep unrolling strategies, this makes the derived models enjoy not only the flexibility and representation capacity of the deep learning approaches but, more importantly, the concise structure of the conventional (optimization-based) methods. Overall, the proposed models constitute a step towards the so-called \textit{interpretable or explainable deep learning}.\\

\section{Problem formulation}\label{problem_formulation}
\begin{figure}
\centering
 \includegraphics[scale=0.35]{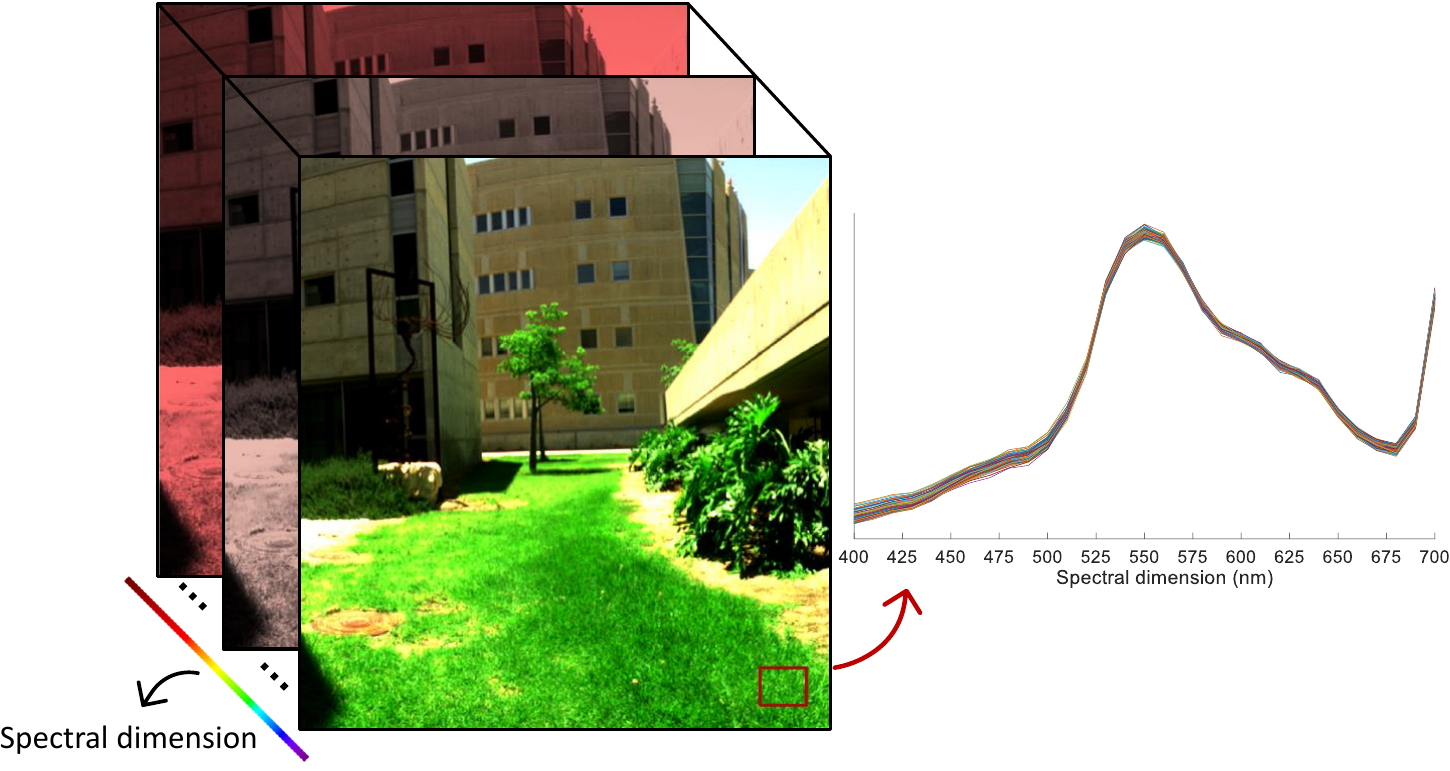}
  \caption{A typical example of signals with strong dependencies across and along (a block of) the data matrix are hyperspectral images, a main property of which is that spatially neighboring hyper-pixels demonstrate strong spectral similarity. In this figure, a small hyperspectral patch with size ($n \times n \times d$) is selected to exemplify this property, thus forming a block of dependent signals $\mat{X} \in \real^{d\times N}$ where $d$ corresponds to spectral dimension of the image and $N=n^2$ is the two stacked spatial dimensions. The dependency \textit{along} the data matrix $\mat{X}$ corresponds to the spectral dimension and the dependency \textit{across} $\mat{X}$  corresponds to the dependencies among the hyper-pixels.   (Image retrieved from \cite{icvl})}.
  \label{fig:example_hyper}
\end{figure}

Consider a data matrix $\mat{Y}'$ that consists of a number of $p$ blocks, $\mat{Y}_i$, $i=1,\ldots,n$ as\begin{equation}
    \mat{Y}'=\left[
    \begin{array}{ccccc}
    \mat{Y}_1&\cdots&\mat{Y}_i&\cdots&\mat{Y}_p
    \end{array}
    \right]\ ,
\end{equation}
where each $\mat{Y}_i \in \real^{d \times N}$ is termed as a block of signals. Consider also that each block of signals is given as
\begin{equation}
    \mat{Y}_i=\mat{X}_i+\mat{W}_i\ ,
\end{equation}
where $\mat{W}_i \in \real^{d \times N}$ denotes a zero-mean noise term and $\mat{X}_i$ 
denotes a \emph{clean} block of signals. In this work we focus on the case where each block $\mat{X}_i$ exhibits dependencies both across and along its dimensions. Particularly, we consider that each column of each block $\mat{X}_i$ has a specific structure (e.g., it admits some sparse representation), and we term this property as dependency \emph{along} the data matrix/block. We also consider that all the columns in each $\mat{X}_i$ have some dependencies, in the sense that the knowledge of a subset of columns gives us some information regarding the other columns, and we term this property as dependency \emph{across} the data matrix/block.  As an example, consider that block $\mat{X}_i$ may correspond to a 
block (patch) of neighboring hyper-pixels of some hyperspectral image. Figure \ref{fig:example_hyper} exemplifies these dependencies. 

In such a setting, given a dictionary $\mat{D} \in \real^{d \times M}$ and focusing on only one block of data, our scope is to compute a sparse representation matrix $\mat{G}_i \in \real ^{M \times N}$, so that
\begin{equation}
    \mat{X}_i \approx \mat{D}\mat{G}_i\ ,
\end{equation}
taking into account the fact that $\mat{X}_i$ exhibits dependencies both along and across its dimensions. In solving this problem, we also utilize knowledge about the assumed structure of $\mat{X}_i$. In this work, we also consider the problem of learning such structure given suitable training data. In the following, since we only focus on the sparse coding of each block of data, we use the symbols $\mat{Y}$ and $\mat{X}$ to refer to one block of data, in order to simplify our notation. At some points, where necessary, we resort to the notation $\mat{Y}_i$ and $\mat{X}_i$ for the blocks of the data, but their use is clear from the context.

\section{Deep architectures for sparse coding} \label{methodA}

Considering the underlying structure of the noisy data block $\mat{Y}$, the studied sparse coding problem can be formulated into the following regularized optimization form
\begin{equation}
   \underset{\mat{G}}{\arg\min}\,\,\, \frac{1}{2} \norm{\mat{Y} - \mat{D}\mat{G}}_F^2 + \mu\norm{\mat{G}}_{1,1} + \lambda \mathcal{R}(\mat{D}\mat{G})\ , 
\label{eq:main_problem}
\end{equation}
that consists of a data consistency term $\norm{\mat{Y} - \mat{D}\mat{G}}_F^2$, a sparsity promoting $l_1$-norm aiming to capture the structure along the data matrix, and some regularization term, denoted as $\mathcal{R}(\cdot)$,  aiming to promote the inherent structure  of the reconstructed
clean estimate $\mat{D}\mat{G}$. 
Furthermore,  $\mu$ and  $\lambda$ are positive scalar constants controlling the relative importance of the sparsity level and the $\mathcal{R}(\cdot)$ prior, respectively. 

To efficiently solve (\ref{eq:main_problem}), an alternating optimization methodology (AO) is employed in order to decouple the data fidelity term, the sparsity term and the regularization term into three individuals sub-problems. To this end, the above optimization problem can be solved efficiently by employing the Half Quadratic Splitting (HQS) methodology \cite{hqs}. By introducing two auxiliary variables, namely $\mat{V}$ and $\mat{Z}$, the problem in (\ref{eq:main_problem}) can be reformulated as follows,
\begin{align}
   \underset{\mat{G}}{\arg\min}\,\,\, &\frac{1}{2} \norm{\mat{Y} - \mat{D}\mat{G}}_F^2 + \mu\norm{\mat{V}}_{1,1} + \lambda \mathcal{R}(\mat{Z}) \\
   &s.t.\quad \mat{V}-\mat{G}=0,\,\, \mat{Z}-\mat{DG}=0. \nonumber
\label{eq:HQSconstr}
\end{align}
The corresponding loss function aiming to solve HQS is given by
\begin{align}
    \mathcal{L}(\mat{G},\mat{V},\mat{Z})= \frac{1}{2} \norm{\mat{Y} - \mat{D}\mat{G}}_F^2 &+ \mu\norm{\mat{V}}_{1,1} + \lambda \mathcal{R}(\mat{Z}) \\ \nonumber   +\frac{b_1}{2}\norm{\mat{V}-\mat{G}}_F^2 
     &+\frac{b_2}{2}\norm{\mat{Z}-\mat{DG}}_F^2, 
\label{eq:lagrangian}
\end{align} where $b_1, b_2>0$ denote a user-deﬁned penalty parameters. Thus, a sequence  of individual subproblems emerges i.e.,\begin{align}
    \mat{G}^{(k+1)} &= \underset{\mat{G}}{\arg\min}\,\,\mathcal{L}(\mat{G}, \mat{V}^{(k)},\mat{Z}^{(k)}) \nonumber\\ 
    \mat{V}^{(k+1)} &= \underset{\mat{V}}{\arg\min}\,\,\mathcal{L}(\mat{G}^{(k+1)}, \mat{V},\mat{Z}^{(k)}) \nonumber\\ 
    \mat{Z}^{(k+1)} &= \underset{\mat{Z}}{\arg\min}\,\,\mathcal{L}(\mat{G}^{(k+1)}, \mat{V}^{(k+1)},\mat{Z}  )
\end{align}
The solutions of the subproblems for $\mat{G}$, $\mat{V}$ and $\mat{Z}$ derive from
\begin{subequations}
\begin{align}
    \mat{G}^{(k+1)} &= (\mat{D}^T\mat{D}+b_2\mat{D}^T\mat{D}+b_1\mat{I})^{-1} \nonumber \\
    &\quad\quad\quad\,(\mat{D}^TY+b_1\mat{V}^{(k)}+b_2\mat{D}^T\mat{Z}^{(k)}) \label{eq:update1} \\
    \mat{V}^{(k+1)} &= soft\,(\mat{G}^{(k+1)},\,\, \mu/b_1) \label{eq:update2} \\
    \mat{Z}^{(k+1)} &=  prox_{\frac{\lambda}{b_2
    }\mathcal{R}}(\mat{D}\mat{G}^{(k+1)}), \label{eq:update3}
\end{align}
\end{subequations}
where the $soft(.,\tau)$ denotes the soft-thresholding operator $x=sign(x)max(\mid{x}\mid-\tau,0)$ and the $prox_h(\cdot)$ stands for the proximal operator of function h. 

Focusing on subproblem in (\ref{eq:update3}), the proximal operator can be written as follows,
\small
\begin{align}
    &prox_{\frac{\lambda}{b_2}\mathcal{R}}(\mat{D}\mat{G}^{(k+1)}) = \nonumber \\
    &\quad\quad\quad\underset{\mat{Z}}{\arg\min} \frac{1}{2(\sqrt{\lambda/b_2})^2}
    \norm{\mat{Z}-\mat{D}\mat{G}^{(k+1)}}_F^2 
    + \mathcal{R}(\mat{Z})
    \label{eq:proximal}
\end{align}
\normalsize
Based on Bayesian estimation theory, relation (\ref{eq:proximal})  can be interpreted as a Gaussian denoiser with noise level $(\sqrt{\lambda/b_2})$. Thus, any Gaussian denoiser in problem (\ref{eq:main_problem}) can be employed to act as a regularizer \cite{Zhangcvpr}, \cite{pmlr-v97-ryu19a}. In the case of interest in this work, we propose replacing  $prox_{\frac{\lambda}{b_2}\mathcal{R}} (\cdot)$ with a  neural network $\mathcal{N}_\theta(\cdot)$ 
, whose weights, denoted as $\theta$,  can be learned from training data. Considering the powerful representation modeling capacity of convolutional neural networks, a CNN denoiser is used   
 to capture the underlying structure priors and the dependencies of the locally dependent signals (e.g., hyperspectral images). \textit{A great merit of the above modeling procedure is the fact that, the explicit prior (regularizer) $\mathcal{R(\cdot)}$ can be unknown in relation (\ref{eq:main_problem}) and be designed with properties adapted to (and trained from) the observed signals.}

Based on the above findings the proposed iterative updated rules of the half quadratic splitting are summarized as follows
\begin{align}\label{eq:finalupdates}
    \mat{G}^{(k+1)} &= (\mat{D}^T\mat{D}+b_2\mat{D}^T\mat{D}+b_1\mat{I})^{-1} \nonumber \\
    &\quad\quad\quad\,(\mat{D}^TY+b_1\mat{V}^{(k)}+b_2\mat{D}^T\mat{Z}^{(k)})\nonumber \\
    \mat{V}^{(k+1)} &= soft\,(\mat{G}^{(k+1)},\,\, \mu/b_1)\nonumber \\
    \mat{Z}^{(k+1)} &= \mathcal{N}_\theta(\mat{D}\mat{G}^{(k+1)}).
\end{align} Note that the neural network $\mathcal{N}_\theta$ is pre-trained according to the following loss function
\begin{equation}\label{eq:loss_pnp}
    \mat{l}(\theta) = \sum_{i=1}^{p} \norm{\mathcal{N}_\theta(\mat{Y}_i;\theta) - \mat{X}_i}_F^2
\end{equation}
where $\theta$ denotes the weights of the CNN prior and $\{\mat{X}_{i}, \mat{Y}_{i}\}$ represent $p$ pairs of the training blocks of locally dependent signals and their corresponding  noisy versions, as for example pairs of ground-truth and their corresponding  noisy hyperspectral image patches. After the training, a simple alternative is to \textit{plug} the learned CNN  into the proposed iterative strategy  defined in (\ref{eq:finalupdates}) and then execute it until convergence is reached. Although, this plug-and-play approach provides empirically sufficient results, the procedure can be considered  piecemeal and sub-optimal, since, in this case, the CNN module is being learnt independently from the forward model (i.e., the dictionary $\mat{D}$) associated with the proposed sparse coding problem. 

\subsection{Deep Unrolling full sparse coding method (DU-full-sc) }\label{DU_A}
\begin{figure*}
\centering
 \includegraphics[width=0.7\linewidth]{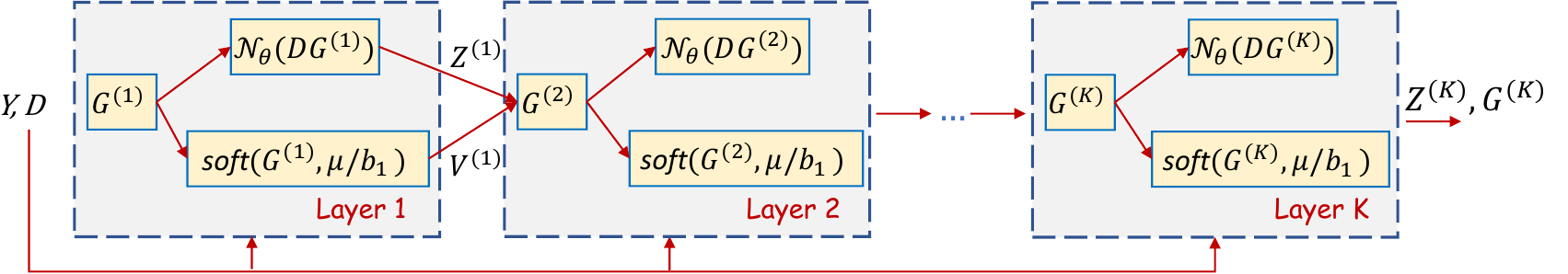}
  \caption{An illustration of our proposed DU-full-sc (Section \ref{DU_A}) model for solving the sparse coding problem in (\ref{eq:main_problem}). Our deep unrolling architecture consists of K layers. Each layer corresponds to a consecutive iteration of the proposed HQS scheme (\ref{eq:finalupdates}). Given the dictionary $\mat{D}$ and a noisy block of  signals $\mat{Y}$ (e.g., noisy hyperspectral image patch), the proposed method aims to generate accurately the denoised version of the block $\mat{Z}^{(K)}$, along with a suitable sparse coding matrix $\mat{G}^{(K)}$ that captures the underlying structure of the data.}
  \label{fig:du_sc}
\end{figure*}

Considering the limitations of the plug-and-play methodology, in this section  a deep unrolling approach is proposed to efficiently tackle these issues by utilizing a form of training called \textit{end-to-end}. In more details, instead of learning the CNN prior $\mathcal{N}_\theta$ offline, we unroll a small number of iterations, say $K$,  of HQS scheme (\ref{eq:finalupdates})  and we treat them as a deep learning network, where each iteration is considered a unique layer of the proposed model. Thus, a K-layer deep learning architecture is formed, where its depth and parameters are  highly interpretable due to the fact that the modeling of the network is based on the physical process underlying the examined problem. Figure \ref{fig:du_sc} illustrates the architecture of a K-layer deep unrolling network.  

By denoting that $\mat{G}^{(K)}$ is the output of the proposed approach, an end-to-end training is performed seeking to minimize some loss function with respect to CNN weights $\theta$. To this end, the following one is employed as loss function,\begin{equation}
   l(\theta) = \sum_{i=1}^{n}{\norm{\mat{D}\mat{G}^{(K)}_i - X_i}^2_F}.
\end{equation}
where $\mat{X}_i$ corresponds to the $i^{th}$ target block of locally dependent signals (e.g., hyperspectral image patch).
Note that the CNN prior $\mathcal{N}_\theta$ is learnt based on the quality of the estimate $\hat{\mat{X}}=\mat{D}\mat{G}^{(K)}$, which strongly depends on the dictionary. Above, it is assumed that all CNN priors $\mathcal{N}_\theta(\cdot)$ of each layer (iteration) have identical weights $\theta$ (weight tying practice), thus reducing the learnable parameters and simplifying the architecture. \textit{Another noteworthy merit of the deep unrolling approach is that  the penalty parameters $\lambda, \mu$ and $b_1, b_2$ introduced in relation  (\ref{eq:HQSconstr}) can be treated also as network parameters to be learned via the end-to-end training scheme.}  
It should be highlighted that an accurate estimate of $\hat{\mat{X}}$  implies that the corresponding sparse coding matrix $\mat{G}^{(K)}$ is enforced to capture the inherent structure of the data.

Note that the number of layers in our deep unrolling approach must be kept  small (5 to 15), This is attributed to the fact that training the deep unrolling network with many layers requires high GPU memory resources, since the calculation of the back-propagation scales with the number of layers \cite{eldar}.  In view of this, in the following section a novel deep equilibrium model is proposed corresponding to an efficient network with infinite number of layers, without demanding high computational resources as the deep unrolling method.

\subsection{Deep equilibrium full sparse coding (DEQ-full-sc)  method}\label{DEQ_A}
Assuming that all instances of the CNN prior $\mathcal{N}_\theta$ have identical weights $\theta$  at each layer of the proposed deep unrolling method, it can be easily observed that the same layer is repeated K times. In other words, the same transformation or iteration map, say $f_\theta$ is applied to the noisy input $\mat{Y}$ in order to obtain an accurate denoised estimate $\hat{\mat{X}}$ of the input and thus an accurate sparse coding matrix $\mat{G}$ capturing the underlying structure of the noisy data.

\begin{figure}
\centering
 \includegraphics[width=1\linewidth]{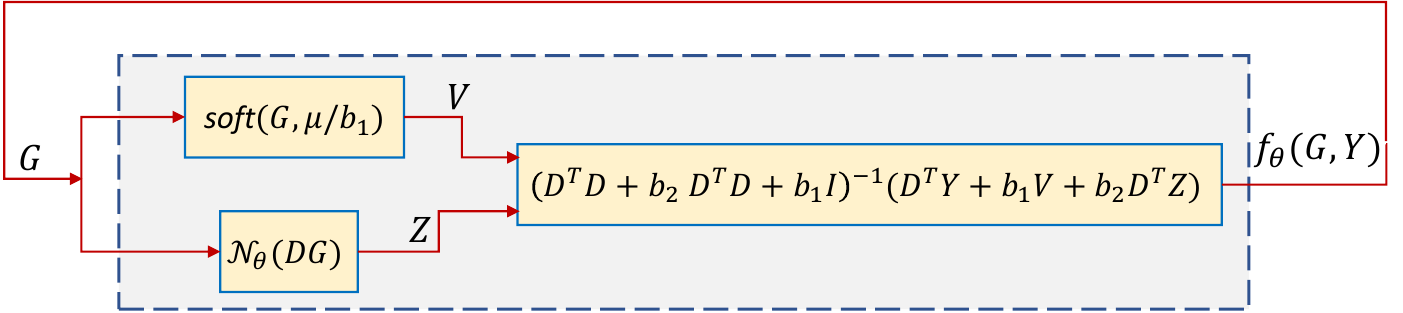}
  \caption{An illustration of the proposed DEQ-full-sc (Section \ref{DEQ_A}) model for sparse coding. 
  The iteration map $f_\theta()$ is designed based on the HQS solver in (\ref{eq:fixed_point_iteration}). Given the dictionary $\mat{D}$ and a noisy block of signals $\mat{Y}$ exhibiting local dependencies along and across the block, the proposed method computes the sparse coding matrix $\mat{G}$, which is the fixed point of the iteration map $f_\theta(\cdot,\mat{Y})$.}
  
  \label{fig:deq_sc}
\end{figure}

In light of this, utilizing the deep equilibrium modeling an implicit infinite-depth network (infinite number of iterations) can be efficiently developed. Thus, the goal is to develop a suitable iteration map $f_\theta(\cdot,y)$ based on the equations in (\ref{eq:finalupdates}). In particular, by substituting the update of $\mat{Z}^{(k+1)}$ and $\mat{V}^{(k+1)}$ directly into the update rule of $\mat{G}^{(k+1)}$, a single and more concrete update expression is derived
\begin{align}\label{eq:fixed_point_iteration}
    &\mat{G}^{(k+1)} = \left(\mat{D}^T\mat{D}+b_2\mat{D}^T\mat{D}+b_1\mat{I}\right)^{-1} \nonumber \\
    &\left(\mat{D}^TY+b_1\,{soft}(\mat{G}^{(k)}, \mu/b_1)+b_2\mat{D}^T\mathcal{N}_\theta(\mat{D}\mat{G}^{(k)})\right).
\end{align} 
Note that the update of $\mat{G}^{(k+1)}$ depends \textbf{only} on the previous iterate $\mat{G}^{(k)}$, thus formulating a fixed point iteration on the sparse coding matrix $\mat{G}$. Hence, the iteration map $f_\theta(\cdot,\mat{Y})$ of relation  (\ref{eq:fixed_point_iteration})  can be written as, 
\begin{equation}\label{eq:iteration_map}
    \mat{G}^{(k+1)} = f_\theta(\mat{G}^{(k)},\mat{Y})
\end{equation}
Based on (\ref{eq:iteration_map}), the estimate of the  target block of locally dependent signals $\mat{X}$, denoted as $\hat{\mat{X}}^\star$ derives from
\begin{equation}
    \hat{\mat{X}}^\star = \mat{D} \mat{G}^\star,
\end{equation}
where $\mat{G}^\star$ is the fixed point of the iteration map $f_\theta(\cdot,Y)$.
Figure \ref{fig:deq_sc} illustrates of the designed iteration map $f_\theta$.

Having effectively designed the iteration map $f_\theta(\cdot,\mat{Y})$, the following challenging procedures emerge. \textbf{First}, given a noisy block of signals $\mat{Y}$ and the weights $\theta$ of the CNN prior, a fixed point is required to be computed during the \textit{forward pass}. \textbf{Second}, given pairs  of ground-truth  blocks of signals exhibiting dependencies along and across the block and their corresponding noisy versions  $\{X_i, Y_i\}_{i=1}^p$, the parameters of the network need to be efficiently obtained during the \textit{training} process.
Without loss of generality, to simplify the notations and calculations below  a single pair of training examples, namely $X,Y$ is considered. Additionally, we use the vectorized versions of the matrices ${Y, X, G}$ denoted as ${y, x , g}$, respectively. 

\subsubsection{Forward pass - Calculating fixed points}\label{forward_pass}
 The "forward" pass during training and inference (testing) in the proposed DEQ model requires computing the fixed point
\begin{equation}
    g^{\star} = f_\theta(g^{\star}, y)
    \label{eq:fixed_point}
\end{equation}
for the iteration map $f_\theta(, y)$ defined in relation (\ref{eq:iteration_map}). A simple strategy to estimate this point is to employ fixed point iterations, i.e., iterating the following recursive scheme
\begin{equation}
    g^{(k+1)} = f_\theta(g^{(k)}, y)
\end{equation}
However, this approach may be a time-demanding procedure. To this end, the Anderson acceleration strategy \cite{aderson} is employed in order to efficiently  accelerate the process of the fixed-point iterations. In particular, instead of using only the previous iterate (i.e., $g^{(k)}$) to compute the next proper point to move, the Anderson acceleration procedure utilizes the m previous iterates formulating the following update rule
\begin{equation}
    g^{(k+1)} = (1-\beta)\sum_{i=0}^{m-1}{\alpha_i g^{(k-i)}} + \beta\sum_{i=0}^{m-1}{\alpha_i f_\theta(g^{(k-i)}, y)},
\end{equation}
for $\beta>0$. The vector $\alpha \in \real^m$ derives from solving the following optimization problem
\begin{equation}
    \arg\min_{\alpha} \norm{U\alpha}^2_2, \quad s.t. \quad 1^T\alpha = 1
    \label{eq:anderson}
\end{equation}
where $U=[f_\theta(g^{k},y)-g^{(k)}, \dots, f_\theta(g^{k-m+1},y)-g^{(k-m+1)} ]$ is a matrix containing the m past residuals. Note that, when m is kept small (e.g., m=5) the computational complexity of optimization problem (\ref{eq:anderson}) is significantly reduced.

\subsubsection{Backward pass - Calculating the Gradient}\label{deq_training}
The implicit backpropagation employed during the training process to obtain the optimal weights $\theta$ of the proposed DEQ model is briefly explained. Based on the findings in \cite{DEQ}, \cite{tutorial}, the goal is to efficiently train the network without requiring to backpropagate though a large number of fixed point iterations.

Let $g^\star=f_\theta(g^\star, y)$ be the fixed point derived from the forward pass and $y$ the noisy block of signals such as a hyperspectral image patch corresponding to the original (noisy-free) patch, denoted as $x$. Let $l(\cdot)$ be a loss function e.g., the mean-squared error (MSE) loss that computes  
\begin{equation}
    l(\theta) = \frac{1}{2}\norm{Dg^\star - x}_2^2
\end{equation}
The loss gradient with respect to the network parameters $\theta$ is 
\begin{equation}
    \frac{\partial l}{\partial \theta} = \frac{\partial g^{\star T} }{\partial \theta} \frac{\partial l}{\partial g^\star} = \left(\frac{\partial g^{\star T} }{\partial \theta}\right)\left( D^T(D g^\star - x)\right),
\label{eq:loss_gradient}
\end{equation}
where the first factor is the Jacobian of $g^\star$ w.r.t. $\theta$ and the second factor is the gradient of loss function. However, in order to compute the first factor, we need to backpropagate through a large number of fixed point iterations. Thus, to effectively avoid this highly demanding task, we employ the following procedure  to calculate the Jacobian of $g^\star$ w.r.t. $\theta$. 

By implicitly differentiating both sides of  fixed point equation (\ref{eq:fixed_point}). i.e., $g^{\star} = f_\theta(g^{\star}, y)$ and applying the multivariate chain rule,  we derive
 \begin{equation}
    \frac{\partial g^\star}{\partial \theta} =
    \frac{\partial f_\theta(g^\star,y)}{\partial g^\star} \frac{\partial g^\star}{\partial \theta} + \frac{\partial f_\theta(g^\star,y)}{\partial \theta} \nonumber
    \label{eq:chain_rule}
    \end{equation}

 Then, solving for $ \frac{\partial g^\star}{\partial \theta}$, an explicit expression for the Jacobian is derived\begin{equation}
    \frac{\partial g^\star}{\partial \theta} = \left(I - \frac{f_\theta(g^\star,y)}{\partial g^\star} \right)^{-1}\frac{f_\theta(g^\star,y)}{\partial \theta} 
    \label{eq:Jacobian}
\end{equation}
Using equation (\ref{eq:Jacobian}), relation (\ref{eq:loss_gradient}) is reformulated as follows
\begin{equation}
    \frac{\partial l}{\partial \theta} = 
     \frac{f_\theta(g^\star,y)^T}{\partial \theta}  \left(I - \frac{f_\theta(g^\star,y)}{\partial g^\star} \right)^{-T} D^T(Dg^\star-x).
\label{eq:gradient2}
\end{equation}
In view of this, instead of backpropagating though a large number of fixed point iterations to compute the loss gradient in relation (\ref{eq:loss_gradient}), we need to calculate only  a memory- and computationally-efficient Jacobian-vector product, as shown in equation (\ref{eq:gradient2}).
Following \cite{tutorial}, in order  to compute this Jacobian-vector product, we  define the  vector $\gamma$ by
\begin{equation}
    \gamma = \left(I - \frac{f_\theta(g^\star,y)}{\partial g^\star} \right)^{-T} D^T(Dg^\star-x)  \nonumber
\end{equation}
which can be rearranged as follows
\begin{equation}
        \gamma = \left(\frac{f_\theta(g^\star,y)}{\partial g^\star} \right)^{T}\gamma + D^T(Dg^\star-x). 
    \label{eq:gamma}
\end{equation}
Note that expression (\ref{eq:gamma}) is also a fixed point equation. Thus, solving this fixed point equation and computing the fixed point $\gamma^\star$, the gradient in 
(\ref{eq:gradient2}) can be written as
\begin{equation}
    \frac{\partial l}{\partial \theta} = 
    \frac{f_\theta(g^\star,y)^T}{\partial \theta} \gamma^\star. 
    \label{eq:grad_loss}
\end{equation}
Thus, given a fixed point  $g^\star$ of an iteration map $f_\theta(g^\star, y)$, the gradient computation procedure is summarized as follows:
\begin{enumerate}
    \item Compute the quantity $D^T(Dg^\star-x)$ .
    \item Compute the fixed point $\gamma^\star$, of equation (\ref{eq:gamma}).
    \item Compute the gradient of the loss function via (\ref{eq:grad_loss}).
\end{enumerate}  
A great benefit from the above analysis is that  the Jacobian-vector product in (\ref{eq:gamma}) can be efficiently computed by conventional automatic differentiation tools.

\section{Deep fast architectures for sparse coding}\label{method B}

The scheme proposed in the previous section successfully captures the dependencies that exist in each block of data, and performs regularized sparse coding at the cost of increased computational complexity. We note that the main reason for this complexity is due to the requirement for a sparse matrix $\mat{G}$, which is promoted via the use of the $l_1$ norm term that is present in the cost function. However, the assumed local dependency
property, suggests that the signal vectors in a block $\mat{Y}$ may have some sort of similarity, as the signals in Figure \ref{fig:example_hyper}. Motivated by this reasoning, in this section we explore a more specific notion of dependency, that suggests that the signal vectors in a block can be represented using the same set of atoms from the dictionary. By adopting such a model, it is possible to drastically reduce the computational complexity of the scheme, since only one set of atoms, i.e., support set, must be determined. In the following we adopt a two step approach, where in the first step a proper support set $\mathcal{S}$ is determined for the whole block of signals $\mat{Y}$, and in the second step, an optimization problem is solved for the computation of the coefficients of the sparse coding matrix, given the support set determined.
At the first step of the proposed approach, the set of atoms that will be used for the representation of the whole block of signals $\mat{Y} \in \real^{d \times N}$ must be determined. For this task, we propose to first compute the average/centroid signal of the block\begin{equation}
    y_c = \frac{1}{N}\sum_{i=1}^N{y_i} \quad \quad y_i \in \real^d, 
\end{equation}
and then employ some sparse coding algorithm (e.g., the OMP) to sparsely encode the vector $y_c$ using the given dictionary. The required support $\mathcal{S}$ is finally given as the set of atoms used in the representation of the vector $y_c$.

Having identified a proper support set $\mathcal{S}$, it is possible to formulate an optimization problem that is significantly less computationally intensive as compared to the problem in (\ref{eq:main_problem}), for the computation of the entries of the sparse coding matrix. In particular, we consider the problem
\begin{equation}
    \arg\min_{G_\mathcal{S}}\norm{\mat{Y} - \mat{D}_\mathcal{S}\mat{G}_\mathcal{S}}^2_F + \lambda \mathcal{R}_\theta(\mat{D}_\mathcal{S}\mat{G}_\mathcal{S})\ ,
    \label{eq:main_problem_2}
\end{equation}
where $\mat{D}_\mathcal{S}$ is the matrix that results from the dictionary $\mat{D}$ after keeping only the columns/atoms indexed by the set $\mathcal{S}$, and $\mat{G}_\mathcal{S}$ is the corresponding matrix of representation weights. It should be noted that matrix $\mat{G}_\mathcal{S}$ is not sparse, since it contains only the nonzero elements of the sparse coding matrix $\mat{G}$.

To sum up, given a block of signals $\mat{Y}$, the following procedure is employed:
\begin{enumerate}
    \item Compute the centroid signal $y_c$ of  the  block $\mat{Y}$.
    \item Determine the support set $\mathcal{S}$ of the centroid signal $y_c$, using some sparse coding algorithm, 
    \item Based on the support of the centroid signal, compute the corresponding representation coefficients for all the signals in $\mat{Y}$ by solving problem (\ref{eq:main_problem_2}).
\end{enumerate}
The details of efficiently solving problem (\ref{eq:main_problem_2})  are provided in the following sub-section.

\subsection{Optimization via HQS}
To efficiently solve the proposed optimization problem in (\ref{eq:main_problem_2}), the HQS methodology is employed again, hence deriving the following constrained optimization form,\begin{align}
   \underset{\mat{G}_\mathcal{S}}{\arg\min}\,\,\, &\frac{1}{2} \norm{\mat{Y} - \mat{D}_\mathcal{S}\mat{G}_\mathcal{S}}_F^2  + \lambda \mathcal{R}(\mat{Z}) \\
   &s.t.\quad  \mat{Z}-\mat{D}_\mathcal{S}\mat{G}_\mathcal{S}=0\ . \nonumber
\label{eq:HQSconstr_2}
\end{align}
The corresponding loss function aiming to solve HQS is\begin{align}
    \mathcal{L}(\mat{G}_\mathcal{S},\mat{Z})= &\frac{1}{2} \norm{\mat{Y} - \mat{D}_\mathcal{S}\mat{G}_\mathcal{S}}_F^2 +  \lambda \mathcal{R}(\mat{Z}) \\ \nonumber   &
     +\frac{b}{2}\norm{\mat{Z}-\mat{D}_\mathcal{S}\mat{G}_\mathcal{S}}_F^2\ , 
\end{align}
where $b>0$ denotes a penalty parameter. 

Similarly with the proposed methods in section \ref{methodA}, we derive the following update rules
\begin{align}
    \mat{G}_\mathcal{S}^{(k+1)} &= (\mat{D}_\mathcal{S}^T\mat{D}_\mathcal{S} + b\mat{D}_\mathcal{S}^T\mat{D}_\mathcal{S})^{-1}\nonumber\\
    &\quad \quad \quad \quad \quad \quad(\mat{D}_\mathcal{S}^TY +b\mat{D}_\mathcal{S}^T\mat{Z}^{(k)})\nonumber\\
    \mat{Z}^{(k+1)} &= prox_{\frac{\lambda}{b}\mathcal{R}}(\mat{D}_\mathcal{S}\mat{G}^{(k+1)}_\mathcal{S})= \mathcal{N}_\theta(\mat{D}_\mathcal{S}\mat{G}^{(k+1)}_\mathcal{S}) 
    \label{eq:hqs_B}
\end{align}

Again the proximal operator $prox_{\frac{\lambda}{b}\mathcal{R}} (\cdot)$ is replaced  with a CNN $\mathcal{N}_\theta(\cdot)$  whose weights can be learned from
training data. 

\subsection{Deep Unrolling fast sparse coding (DU-fast-sc)  method}\label{DU_B}
\begin{figure*}
\centering
 \includegraphics[scale=0.5]{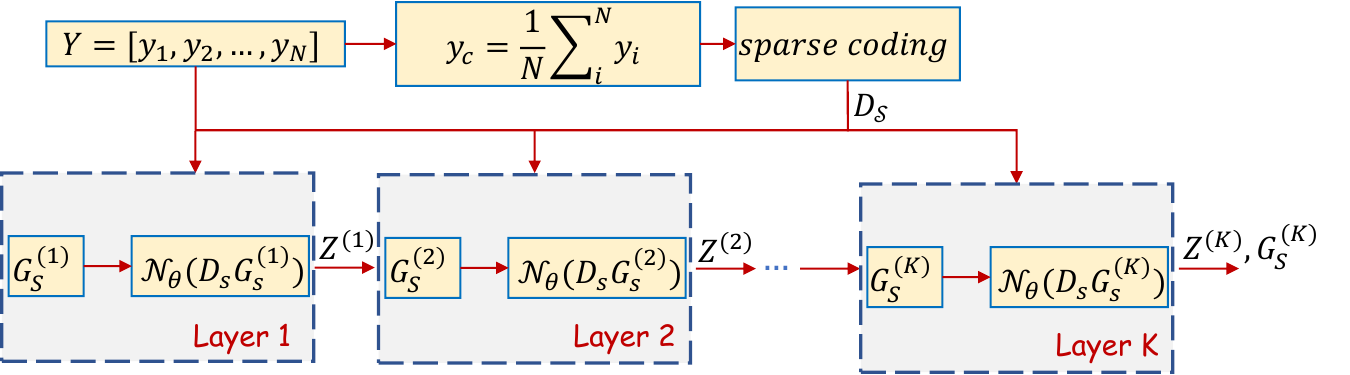}
  \caption{An illustration of 
  the proposed DU-fast-sc (Section \ref{DU_B}) for solving the sparse coding problem, for one block $\mat{Y}$. The proposed architecture consists of two main stages. First, based on the centroid signal of the block Y the most suitable atoms from the dictionary, denoted as $\mat{D}_\mathcal{S}$ are selected to represent the signals in $\mat{Y}$. After that, the proposed Deep Unrolling method is deployed in order to accurately compute the corresponding representation coefficients $\mat{G}_\mathcal{S}$ for each signal in the block $\mat{Y}$. }
  \label{fig:DU_fast}
\end{figure*}

Similar to the Deep Unrolling method developed in Section \ref{DU_A}, a small number of iterations ($K$) of the HQS scheme in (\ref{eq:hqs_B}) can be unrolled, thus forming a $K$-layer deep learning model. After that, the learnable parameters of the proposed architecture, namely the CNN denoiser $\mathcal{N}_\theta(\cdot)$ and the penalty parameters $\lambda$ and $b$, can be learnt via end-to-end training.  
Given pairs of noisy/corrupted and corresponding ground-truth blocks of signals $\{\mat{Y}_{i}, \mat{X}_{i}\}$, we seek to minimize the following loss function $  l(\theta) = \sum_{i=1}^{n}{\norm{\mat{D}_{\mathcal{S}_i}\mat{G}^{(K)}_{\mathcal{S}_i} - X_i}^2_F},$
where $\mat{D}_{\mathcal{S}_i}$ corresponds to the selected atoms derived from the centroid signal of the $i^{th}$ noisy block $\mat{Y}_i$ and $\mat{G}^{(K)}_{\mathcal{S}_i}$ is the output of our deep unrolling method. Note that alternatively we can use directly the $\mat{Z}_i^{(K)}$ as an accurate estimate of the target block $\mat{X}_i$. Figure \ref{fig:DU_fast} provides an illustration of the proposed deep unrolling method.

\subsection{Deep equilibrium fast sparse coding (DEQ-fast-sc) method}\label{DEQ_B}

Following the methodology in Section \ref{DEQ_A}, our goal is to design again an efficient iteration map $f_\theta(\cdot,Y)$ utilizing the equations in (\ref{eq:hqs_B}). To simplify further the update rules, we substitute the expression for $\mat{Z}^{(k+1)}$ into the expression for $\mat{G}^{(k+1)}$, thus obtaining the following update rule\begin{align}
    \mat{G}_\mathcal{S}^{(k+1)} &= \left(\mat{D}_\mathcal{S}^T\mat{D}_\mathcal{S} + b\mat{D}_\mathcal{S}^T\mat{D}_\mathcal{S}\right)^{-1}\nonumber\\
    &\quad \quad \quad \quad \quad\left(\mat{D}_\mathcal{S}^TY +b\mat{D}_\mathcal{S}^T\mathcal{N}_\theta(\mat{D}_\mathcal{S}\mat{G}^{(k)}_\mathcal{S})\right)
    \label{eq:fixed_point_iteration_B}
\end{align}
Thus, equation (\ref{eq:fixed_point_iteration_B}) can be considered as a fixed point iteration of the variable $\mat{G}_\mathcal{S}$, where the iteration map satisfies the following recursive scheme
\begin{equation}
    \mat{G}_\mathcal{S}^{(k+1)} = f_\theta(\mat{G}_\mathcal{S}^{(k)},\mat{Y})
\end{equation}

\begin{figure}
\centering
 \includegraphics[width=1\linewidth]{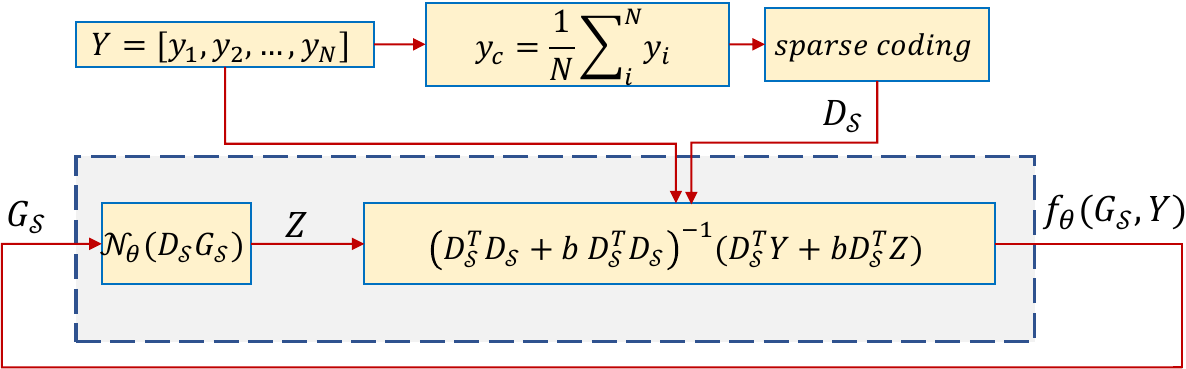}
  \caption{An illustration of 
  the proposed DEQ-fast-sc approach (Section \ref{DEQ_B}) for solving the sparse coding problem, for one block $\mat{Y}$. The proposed architecture consists of two main stages. First, based on the centroid signal of the block Y the most suitable atoms from the dictionary, denoted as $\mat{D}_\mathcal{S}$ are selected to represent the signals in  $\mat{Y}$. After that, the proposed Deep Equilibrium method is deployed in order to accurately compute the corresponding representation coefficients $\mat{G}_\mathcal{S}$ for each signal in the block $\mat{Y}$. The  output matrix $\mat{G}_\mathcal{S}$ is the fixed point of the iteration map $f_\theta(\cdot,\mat{Y})$ defined in (\ref{eq:fixed_point_iteration_B}). 
}
  \label{fig:deq_fast}
\end{figure}

The corresponding  estimate of the target block of locally dependent signals $\mat{X}$, denoted as $Z^\star$ is given by
\begin{equation}
    \mat{Z}^\star = \mat{D}_{\mathcal{S}} \mat{G}_\mathcal{S}^\star,
\end{equation}
where $\mat{G}^\star$ is the fixed point of the iteration map $f_\theta(\cdot,Y)$.
Figure \ref{fig:deq_fast} provides an illustration of the proposed DEQ approach.
Details concerning the calculation of fixed points and the training procedure are given in the previous Sections \ref{forward_pass} and \ref{deq_training}, accordingly. 

\section{Experimental part}\label{Results}
In this section, extensive numerical experiments are presented, in the context of hyperspectral imaging, to validate the efficacy and applicability of the proposed sparse coding schemes. In more detail, the problem of hyperspectral image (HSI) denoising is considered, where the proposed deep unrolling  and deep equilibrium models are compared with  various sparse coding algorithms and state-of-the-art HSI denoising approaches. As it will be shown, based on solid experimental results, the proposed sparse coding approaches:
\begin{itemize}
    \item Perform remarkably better than classical sparse coding based algorithms.
    \item Exhibit superior performance compared to  state-of-the-art deep learning based techniques.
        \item Notably outperform the plug-and-play approaches.
\end{itemize}

\subsection{Dataset}
To demonstrate the merits of the proposed models, we used a publicly available hyperspectral image dataset, pertaining to a variety of natural scenes. In more detail, hyperspectral images from the iCVL dataset \cite{icvl} were employed to train and validate our models. From this dataset, 70 hyperspectral images were employed to generate the training set, whereas another 50 images  were used to generate the testing set. The examined hyperspectral images constitute  $1300\times1300\times31$ dimensional cubes, where the last dimension $d=31$ corresponds to the spectral dimension (i.e., $31$ spectral bands in the $400-700$ nm spectrum). 

\subsection{Implementation Details}

\subsubsection{Block processing}
\begin{figure*}
\centering
  \subfloat{
    \includegraphics[scale=0.42]{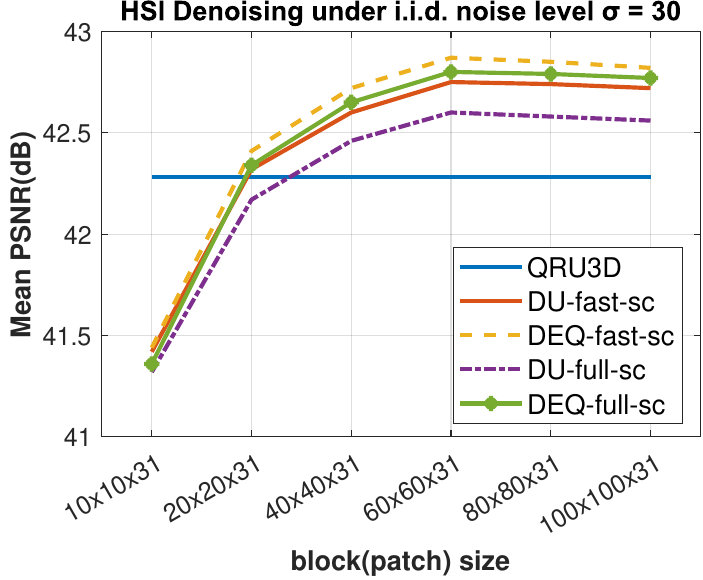}}
   \hfill \null
  \subfloat{
    \includegraphics[scale=0.42]{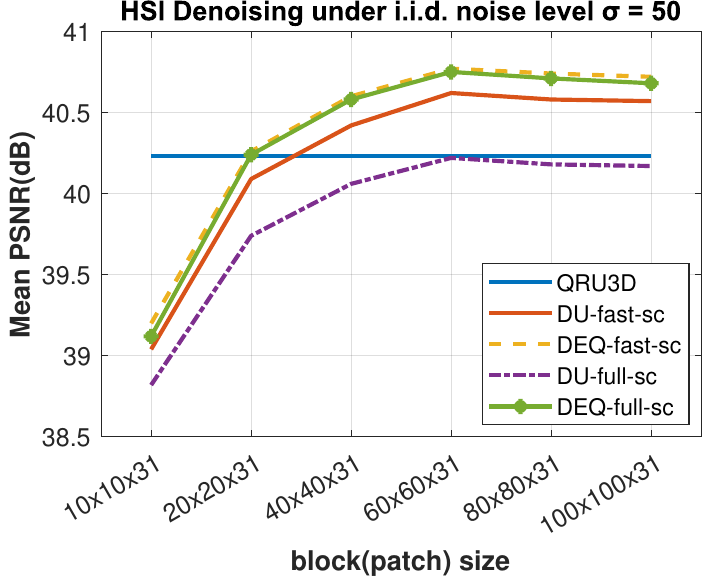}}
    \hfill\null
  \subfloat{
    \includegraphics[scale=0.42]{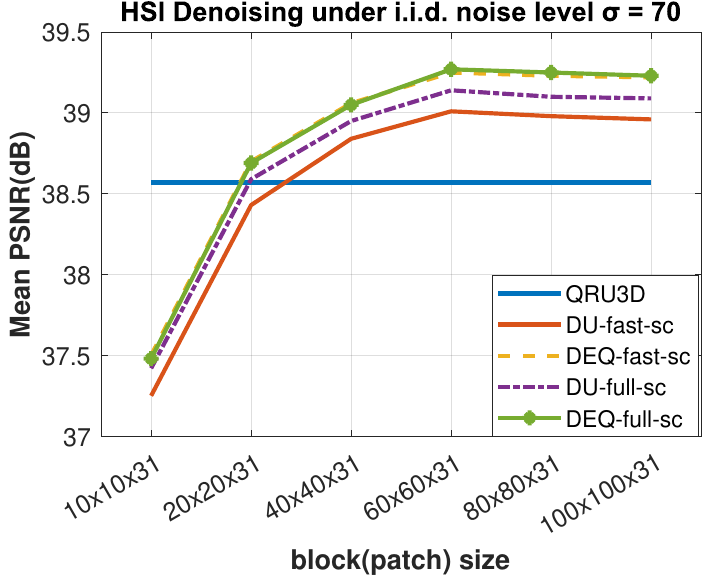}}
  \caption{Influence of the block(patch) size ($n\times n\times 31$) on the reconstruction accuracy of the proposed models. Note that that the third dimension (i.e., 31)  corresponds to the number of spectral bands of the hyperspectral images. Additionally, the performance of the state-of-the-art model QRU3D \cite{QRU3D} is also depicted to highlight that the proposed models are able to maintain competitive performance for a wide range of block size values.}
  \label{fig:patch_noise}
\end{figure*}

The training as well as the testing datasets used in our experiments contain \emph{blocks (patches)} collected by the hyperspectral images. Each such block (patch) has a size of $n\times n \times d$. The parameter $n$, that defines the spatial size of each block (patch), plays an important role in the accuracy as well as in the computational complexity of the proposed approaches. In order to obtain a proper value for the parameter $n$, several experiments with different values for $n$ and noise levels were conducted. Figure \ref{fig:patch_noise} presents the resulting hyperspectral image denoising performance, in terms of the Peak Signal to Noise Ratio (PSNR), for various values of the parameter $n$. As it can be seen from these results, the value $n=60$ gave the best performance, for the considered dataset. Thus, for the rest of our experiments the block(patch) size was set equal to $60\times60\times31$. allowing the proposed methods to exhibit remarkably low computational complexity (and associated  runtimes), as can be seen from Table \ref{tab:runtime} 

\subsubsection{Training and Testing setting} During the \textbf{training phase}, the hyperspectral images are corrupted by several types of noise.
After that we split  both ground-truth and noisy images  into non-overlapping patches(blocks) as described above and then we randomly sample  pairs of hyperspectral patches (i.e., blocks of  signals, which exhibit dependencies along and across the block ) and their corresponding noisy versions, denoted as $\{X_i, Y_i\}_{i=1}^p$. 
During the \textbf{testing phase},  the following procedure was employed. Initially, the image was corrupted  with a specified type of noise, and then it was processed into $n\times n \times d$ non-overlapping patches. For each noisy block, say $\mat{Y}$ the corresponding sparse coding matrix was computed using the proposed schemes. 
To validate the accuracy of the sparse coding models we compared the reconstructed HSI with the corresponding original image in terms of the Peak Signal to Noise Ratio (PSNR), the 
Structural Similarity Index (SSIM) \cite{sim} metrics and the Spectral Angle Mapper (SAM) \cite{sam}. 
\subsubsection{Noise setting}\label{noise_set} Real-world hyperspectral images are corrupted by several types of noise e.g., Gaussian noise, dead pixels or lines, and stripes \cite{QRU3D, Rasti_review}. To this end, following the experimental set up of the study in \cite{QRU3D}, we examine two noise scenarios. In the first scenario, the images are corrupted by i.i.d. Gaussian noise corresponding to three different noise levels with $\sigma = 30, 50, 70$, referred to as i.i.d Gaussian noise.

In the second scenario, a more realistic case is explored considering three types of mixed noise types:\begin{itemize}
\item Case 1: Non-i.i.d. Gaussian noise. The entries of the different spectral bands are corrupted by zero-mean Gaussian noise with different noise intensities, in particular, corresponding to $\sigma$ randomly selected from 10 to 70.
\item Case 2: Non-i.i.d. Gaussian + Stripe noise. All bands are corrupted by non-i.i.d. Gaussian noise as in Case 1. One third of the bands (10 bands) are randomly chosen to add stripe noise (5 to 15 percentages of columns) \cite{QRU3D}.
\item Case 3: Non-i.i.d. Gaussian + Deadline noise. The noise generation process is similar as in Case 2 except that the stripe noise is replaced by a so-called deadline noise \cite{QRU3D}.
\end{itemize}
\subsubsection{CNN architecture}
In our proposed models, a relatively small CNN module $\mathcal{N}_\theta(\cdot)$ was employed with 4 layers. In more detail, each convolutional layer comprised of 64 filters with size $3 \times 3$ followed by a non-linear activation function ReLU. Following the remarks of \cite{pmlr-v97-ryu19a} the spectral normalization \cite{spectral_norm} was imposed to all layers, thus guaranteeing that each layer has Lipschitz constant (no more than) 1. The Regularizing Lipschitz continuity was employed not only during the pre-training of the CNN network but also during the training phase of the proposed sparse coding models, hence providing training stabilization and improved performance.

\subsubsection{Dictionary}\label{sec_dict}
The analysis that has been conducted assumes that the dictionary $\mat{D} \in \real^{d\times M}$ ($d=31$) is known. Thus, in all experiments a fixed  dictionary was employed that had been learnt using the KSVD \cite{Elad2010}  and OMP algorithms over a collection of noisy hyperspectral images from the training set and different from the images considered during the testing phase. The considered dictionary contained $M=512$ atoms.

\subsubsection{Hyperparameters setting}
Based on the training dataset, the CNN module $\mathcal{N}_\theta(\cdot)$ was pre-trained utilizing the ADAM optimizer via loss function (\ref{eq:loss_pnp}). In particular, the number of epochs was set to 150, the learning rate was set to $1e-03$ and the batch size was 16.
Subsequently, this pre-trained version of 
$\mathcal{N}_\theta$ is used for initializing the respective model of the proposed deep unrolling and deep equilibrium architectures, and is further trained during the end-to-end training procedure.   

Regarding the deep unrolling methods described in Section \ref{DU_A} and \ref{DU_B} (i.e., DU-full-sc and DU-fast-sc) the number of unrolling iterations (or the number of the layers) was set to $K=10$. To proceed further, concerning the deep equilibrium methods defined in Section \ref{DEQ_A} and \ref{DEQ_B} (i.e., DEQ-full-sc and DEQ-fast-sc) the Anderson acceleration procedure \cite{aderson} was deployed during the training phase for the forward and backward pass fixed-point iterations. Specifically, the number of fixed-point iteration during the forward and backward pass  was set to 20. Finally, during the end-to-end training phase of the deep unrolling and deep equilibrium models, the ADAM optimizer was deployed to update the network parameters. In this case, the number of epochs was set to 100, the learning rate was set to $1e-04$ and the batch size to $16$. \textit{Note that a strong advantage of the proposed models is the fact that all the penalty parameters in equation (\ref{eq:finalupdates}) can be treated also as network
parameters to be learned via the end-to-end training.}

\subsection{Hyperspectral Denoising performance}
\subsubsection{Proposed sparse coding models against various sparse coding algorithms}
Since the proposed sparse representation deep unrolling and equilibrium models are essentially generic sparse coding algorithms, in this section, a thorough comparison is performed with several sparse coding algorithms.
Table \ref{tab:results_sparse} summarizes the average  quantitative results of our proposed methods, namely the DU-full-sc (Section \ref{DU_A}), the DEQ-full-sc (Section \ref{DEQ_A}), the DU-fast-sc (Section \ref{DU_B}) and the DEQ-fast-sc (Section \ref{DEQ_B}) in comparison with various well-established sparse coding algorithms. It is evident that the proposed approaches are markedly
 better than the other sparse coding methods under comparison. 
 Since the hyperspectral images are a great example of signals containing blocks with strong dependencies, the incorporation of a learnable regularizer (CNN module) and the transformation of the sparse coding  optimization schemes into a meaningful and highly interpretable deep learning architectures enable the proposed methods to model effectively the underlying structure of the noisy signals, thus offering great denoising properties. 

\begin{table*}
  \centering
  \caption{Proposed methods versus various sparse coding approaches: results under several i.i.d. Gaussian noise levels on ICVL dataset.}
  
   \resizebox{15cm}{!}{
  \begin{tabular}{ccccccccccl}
    \toprule
    Sigma& Metrics&Noisy  & batch-OMP  & Lasso  & SunSaL-TV  & fast-TV  &DU-full-sc & DEQ-full-sc &DU-fast-sc &DEQ-fast-sc \\
    & & &\cite{batch_omp} & \cite{lasso} & \cite{sunsanltv} & \cite{mypaper} & Section \ref{DU_A} & Section \ref{DEQ_A} & Section \ref{DU_B} &Section \ref{DEQ_B}\\ 
    
    \midrule
         &PSNR& 18.54 & 36.03 & 36.41   &37.98   & 38.62   & 42.60  & 42.80  & 42.75  & \textbf{42.87} \\
      30 &SSIM& 0.112 & 0.889 & 0.917   & 0.939 & 0.946   & 0.972  & 0.973  & 0.974  & \textbf{0.975} \\
         &SAM&  0.807 & 0.146 &  0.113  & 0.089  & 0.082   & 0.045  & 0.042  & 0.042  & \textbf{0.041}\\
    \midrule
         &PSNR& 14.12 & 33.17 & 33.54   & 36.13 & 36.74  & 40.22  & 40.75  & 40.62  & \textbf{40.77} \\
      50 &SSIM& 0.043 & 0.751 & 0.849   & 0.927 & 0.908  & 0.956  & 0.962  & 0.961  & \textbf{0.963} \\
         &SAM& 0.993  & 0.232 & 0.098   &0.101  &  0.093  & 0.054  & \textbf{0.047}  & 0.050   & 0.048\\
    \midrule
         &PSNR& 11.21 &30.91  &31.30  &34.49  & 35.02   & 39.14  & \textbf{39.27}  & 39.01  & 39.25 \\
      70 &SSIM& 0.024 &0.660  &0.759  &0.869  & 0.875   & 0.947  & 0.949  & 0.948  & \textbf{0.950} \\
         &SAM&  1.102 &0.280  &0.130  &0.126  & 0.118   & 0.056  & \textbf{0.054} & 0.056  & 0.055\\
    \bottomrule
  \end{tabular}}
    \label{tab:results_sparse}
\end{table*}

\subsubsection{Proposed sparse coding models against state-of-the-art Hyperspectral denoising methods} 

\textbf{Competing methods.}
To fully demonstrate the merits and potentials of the proposed models,  we compare our methods with several representative conventional as well as deep learning based approaches for both i.i.d. Gaussian and mixed non-i.i.d. noise cases.

Following a methodology similar to \cite{QRU3D}, we note that most conventional approaches are more suitable to tackle noise cases with specific characteristics compatible to their noise assumption. On the other hand, deep-learning based approaches are able to tackle various noise scenarios. In order to perform a fair comparison, we employed different well studied conventional methods in the two considered noise scenarios (Section \ref{noise_set}).

To this end, in the i.i.d. Gaussian noise scenario, we compare  with some optimization-based/baselines approaches including ﬁltering-based approaches (BM4D \cite{bm4d}), dictionary learning (TDL \cite{TDL}) and tensor-based  (ITSReg \cite{itsreg}, LLRT \cite{LLRT}) approaches. In the mixed
non-i.i.d noise scenario, we adopt several classical optimization-based methodologies, including low-
rank matrix recovery approaches (\cite{LRMR, LRTV, NMoG, fasthyde, hyres}), and a low-rank tensor approach (\cite{TDTV}).
Concerning the deep learning approaches, we consider the state-of-the-art  QRU3D \cite{QRU3D} model, and also the MemNet \cite{MemNet}, and the  HSID-CNN \cite{HSID-CNN} approaches. 
Note that we have carefully replicated all the simulation parameters and have used the exact same dataset (training and testing images) as used in \cite{QRU3D}, thus ensuring a fair comparison with the results appearing in that study. Regrading, the FastHyde method \cite{fasthyde}, we use the Hyperspectral Denoising toolbox, called HyDE \footnote{https://github.com/Helmholtz-AI-Energy/HyDe} \cite{Hyde}. 
 
\textbf{Denoising in the i.i.d. Gaussian Noise Case:}
In this noise scenario, the hyperspectral images are corrupted with zero mean i.i.d Gaussian noise with  three different
noise levels (more details in Section \ref{noise_set}). 
Table \ref{tab:results_deep} summarizes the main reconstruction accuracy comparison. As can be clearly seen, the proposed Deep Unrolling and especially the Deep Equilibrium-based models exhibit better performance results as compared to the other approaches. Although the proposed models are not designed explicitly for the HSI denoising problem, their modeling capacity based on the sparsity  and the CNN-learnable regularizer enable them to provide competitive performance against both traditional and deep learning methodologies designed to tackle only the  examined problem.

\textbf{Denoising in a Mixed Noise Cases:}
To strengthen the experiments, and  demonstrate the capabilities of the proposed models, a more realistic scenario is considered using three types of mixed noise to generate the noisy samples. In brief, cases 1-3 represent non-i.i.d Gaussian noise,  non-i.i.d Gaussian + stripes, non-i.i.d  Gaussian + deadline (see Section \ref{noise_set}). Table \ref{tab:results_deep_complex} summarizes the results. In more detail, the proposed models demonstrate competitive performance against the state-of-the-art model QRU3D and notably outperform the other approaches. Furthermore, the proposed deep equilibrium models and especially the fast version (i.e., DEQ-fast-sc) provide better reconstruction results compared to the QRU3D method for the task of removing mixed types of non i.i.d noise. 

\begin{figure}
\centering
\includegraphics[scale=0.34]{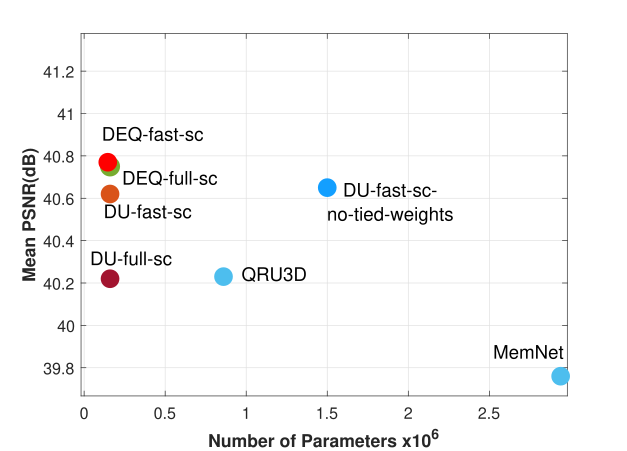}
     \caption{Model complexities comparison of our proposed sparse coding schemes and two state-of-the-art networks under the i.i.d Gaussian noise scenario with $\sigma=50$. Although, the proposed models require $82.5\%$ and $95\%$ less parameters compared to the QRU3D and MemNet models, respectively, they are able to provide better or competitive performance. Additionally, for the DU-fast-sc method we consider also the case where the CNN network is different at each layer (i.e., the DU-fast-sc-non-tied-weights.) 
}
  \label{fig:num_of_param}
\end{figure}

Figure \ref{fig:num_of_param} sheds light on the above results by presenting the number of learnable parameters  of the models, which are involved in this experiment. It is notable that our proposed methods not only achieve competitive performance but also require considerably less  parameters as compared to the two best performing  deep learning-based approaches, i.e., the state-of-the-art model QRU3D and MemNet. In more detail,  the proposed models require \textbf{$\mathbf{82.5\%}$ and $\mathbf{95\%}$ less parameters} as compared to the QRU3D and MemNet models. The above interesting  remark can be justified by taking into account that the proposed models have well-justified architectures derived from  modeling of the underlying physical processes  and  utilizing  prior domain knowledge, in the form of correlation structure and sparsity priors. Thus, the proposed methods enjoy both the modeling capacity of the deep-learning methods, and the concise structure of the sparse coding algorithms.

\begin{table*}
  \caption{Proposed methods versus several state-of-the-art denoising methods: Results under several i.i.d. Gaussian noise levels on ICVL dataset.}
  \resizebox{\linewidth}{!}{
  \begin{tabular}{ccccccccccccccl}
    \toprule
    i.i.d. noise& Metrics&Noisy &BM4D &ITSReg&FastHyde$^1$ & TDL  & LLRT  & HSID-CNN  & MemNet  & QRU3D   &DU-full-sc & DEQ-full-sc &DU-fast-sc &DEQ-fast-sc \\
    sigma& & &\cite{bm4d}&\cite{itsreg}&\cite{fasthyde} &\cite{TDL} & \cite{LLRT} &  \cite{HSID-CNN} &  \cite{MemNet}   &\cite{QRU3D} & Section \ref{DU_A} & Section \ref{DEQ_A} & Section \ref{DU_B} &Section \ref{DEQ_B}\\ 
    \midrule
         &PSNR& 18.54& 38.45 & 41.48 &38.37  & 40.58 & 41.99 & 38.70 & 41.45 & 42.28 & 42.60  & 42.80  & 42.75  & \textbf{42.87} \\
      30 &SSIM& 0.112& 0.934 & 0.961 &0.951  &0.957 & 0.967 & 0.949 & 0.972 & 0.973 & 0.972  & 0.973  & 0.974  & \textbf{0.975} \\
         &SAM&  0.807& 0.126 & 0.088 &0.090  &0.062 & 0.056 & 0.103 & 0.065 & 0.061 & 0.045  & 0.042  & 0.042  & \textbf{0.041}\\
    \midrule
         &PSNR& 14.12&35.60 &38.88 &37.42  &38.01 & 38.99 & 36.17 & 39.76 & 40.23 & 40.22  & 40.75  & 40.62  & \textbf{40.77} \\
      50 &SSIM& 0.043&0.889 &0.941 &0.946  &0.932 & 0.945 & 0.919 & 0.960 & 0.961 & 0.956  & 0.962  & 0.961  & \textbf{0.963} \\
         &SAM& 0.993&0.169  &0.098 &0.095  &0.085 & 0.075 & 0.134 & 0.076 & 0.072 & 0.054  & \textbf{0.047}  & 0.050   & 0.048\\
    \midrule
         &PSNR& 11.21&33.70 &36.71&36.14  &36.36 & 37.36 & 34.31 & 38.37 & 38.57 & 39.14  & \textbf{39.27}  & 39.01  & 39.25 \\
      70 &SSIM& 0.024&0.845 &0.923&0.938  &0.909 & 0.930 & 0.886 & 0.946 & 0.945 & 0.947  & 0.949  & 0.948  & \textbf{0.950} \\
         &SAM&  1.102&0.207 &0.112&0.099   &0.105 & 0.087 & 0.161 & 0.088 & 0.087 & 0.056  & \textbf{0.054} & 0.056  & 0.055\\
    \bottomrule
  \end{tabular}}
    \label{tab:results_deep}
\end{table*}

\begin{table*}
  \caption{Proposed methods versus several state-of-the-art denoising methods: Results under  Mixed noise cases on ICVL dataset.}
  \resizebox{\linewidth}{!}{
  \begin{tabular}{ccccccccccccccl}
    \toprule
    & Metrics&Noisy &LRMR &LRTV & NMoG  & TDTV& FastHyde$^1$  & HSID-CNN  & MemNet & QRU3D   &DU-full-sc & DEQ-full-sc &DU-fast-sc &DEQ-fast-sc \\
    noise& & &\cite{LRMR}&\cite{LRTV} &\cite{NMoG} & \cite{TDTV} & \cite{fasthyde} &\cite{HSID-CNN} &  \cite{MemNet}   &\cite{QRU3D} & Section \ref{DU_A} & Section \ref{DEQ_A} & Section \ref{DU_B} &Section \ref{DEQ_B}\\ 
    \midrule
     Case 1    &PSNR    & 18.24  & 32.80 & 33.62 & 34.51 & 38.14& 38.83  & 38.40 & 38.94  & 42.79          & 42.31  & 42.84     & 42.63      & \textbf{42.91} \\
      non i.i.d.         &SSIM& 0.168  & 0.719 & 0.905 & 0.812 & 0.944& 0.947   & 0.947 & 0.949   & \textbf{0.978} & 0.969  & 0.974      & 0.970      & 0.976 \\
               &SAM     &  0.897 & 0.185 & 0.077 & 0.187 & 0.075&0.125    & 0.095 & 0.091   & 0.052          & 0.053  & 0.051      & 0.051      & \textbf{0.049}\\
    \midrule
      Case 2   &PSNR& 17.80      &32.62  &33.49 & 33.87  & 37.67& 38.28   & 37.77 & 38.57  & 42.35 & 42.15   & 42.48           & 42.32           &  \textbf{42.55}\\
     non i.i.d.  &SSIM& 0.159  &0.717  &0.905 &0.799   & 0.940&0.936    & 0.942 & 0.945  & 0.976 & 0.965   & 0.975           & 0.970            &  \textbf{0.977}\\
     + stripes     &SAM& 0.910      &0.187  &0.078 &0.265   & 0.081&0.142   & 0.104  & 0.095  & 0.055 & 0.058   & 0.052           & 0.055            & \textbf{0.052}\\
    \midrule
      Case 3    &PSNR& 17.60      &31.83  &32.37 & 32.87  & 36.15&37.31   & 37.65 & 38.15  & 42.23 & 42.18   & 42.35           & 42.25        & \textbf{42.44} \\
    non i.i.d.  &SSIM& 0.155 &0.709  &0.895 &0.797  & 0.930&0.918   & 0.940 & 0.945  & 0.976 & 0.961   & 0.973             & 0.0971        &  \textbf{0.975}\\
    + deadline     &SAM& 0.917       &0.227  &0.115 &0.276   & 0.099&0.106   & 0.102  & 0.096  & 0.056 & 0.060 & 0.055      & 0.058         & \textbf{0.055}\\
    \bottomrule
  \end{tabular}}
    \label{tab:results_deep_complex}
\end{table*}

\textbf{Denoising in the high SNR regime:}
Considering that in real-world applications the SNR of the acquired hyperspectral images is quite high, typically in the range 20 to 50 dB, in this section, we compared   the proposed two best-performing deep equilibrium models against the state-of-the-art model QRU3D and the MemNet model and some best-performing conventional approaches in the case where the SNR is equal to 40dB. Since the conventional methods perform better in high SNR values, in this experiment, we have included some additional methods that we omitted in the previous experiments.
Table \ref{tab:high_psnr} summarizes the results. In the case of high SNR values the conventional approaches  provide better reconstruction results as compared to  the deep learning architectures. However, the proposed deep equilibrium models are able to exhibit superior performance against both the conventional and the state-of-the-art QRU3D model. The above remark can be justified by the fact that the proposed equilibrium models have been derived through the transformation of the sparse
coding optimization schemes into a meaningful and highly
interpretable deep learning architectures, which enable the proposed
methods to model effectively and accurately the examined problem.

\begin{table*}
\centering
  \caption{Proposed methods versus several denoising methods under high SNR regime.}

  \resizebox{15cm}{!}{
  \begin{tabular}{ccccccccccl}
    \toprule
    Noise& Metrics&Noisy&ItsReg &LLRT & FastHyde$^{1}$       & FORDN$^{1}$   &Hyres$^{1}$          &QRU3D          & DEQ-full-sc & DEQ-fast-sc\\
         &        &     &\cite{itsreg}&\cite{LLRT}&\cite{fasthyde}&\cite{fordn}& \cite{hyres}            &\cite{QRU3D} &                 &           \\               
    \midrule
     i.i.d.&PSNR  &40.5&50.05&50.17 &51.16   & 49.25&48.81  & 48.85  & 52.26             & \textbf{52.35}             \\
    \bottomrule
  \end{tabular}}
    \label{tab:high_psnr}
\end{table*}

\subsubsection{Proposed sparse coding models against the Plug-and-play approaches}
In this section a thorough comparison is provided between the proposed deep unrolling and deep equilibrium models, on the hand, and the corresponding plug-and-play (PnP) approaches, termed as PnP-full-sc and PnP-fast-sc respectively. Note that the PnP methodologies can be derived from the iterative solvers in (\ref{eq:finalupdates}) and (\ref{eq:hqs_B}), where a pre-trained  neural network is  plugged into the  iterative algorithms and execute them until
convergence is reached. According to Table \ref{tab:results_pnp}, the proposed models remarkably outperform the plug-and-play approaches. This stems from the fact that the plug-and-play methods are not optimized end-to-end, thus the CNN
network is trained independently form the considered problem and the forward
model i.e., the dictionary.

\begin{figure}
\centering
 \includegraphics[scale=0.42]{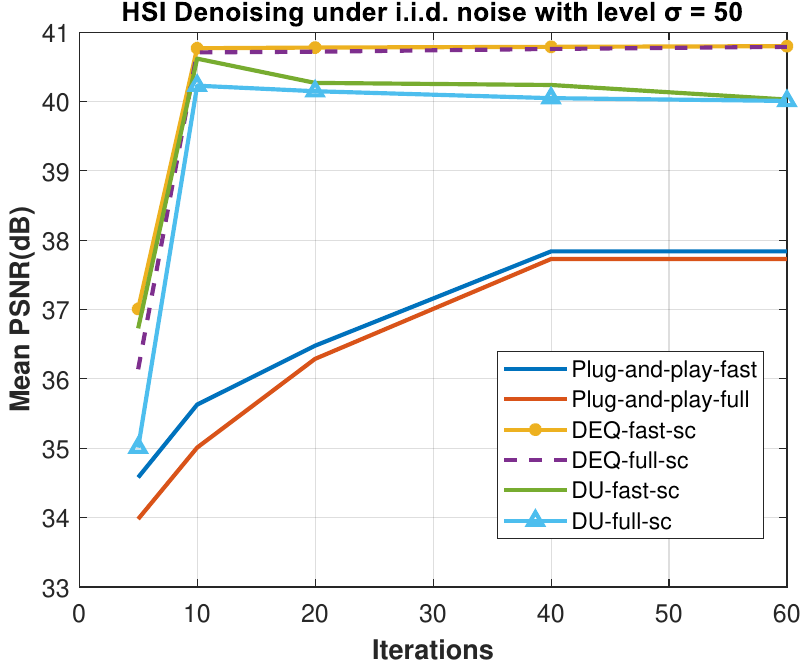}
  \caption{Iterations vs PSNR of reconstructed images  for the proposed deep unrolling (i.e, DU-full-sc, DU-fast-sc) and deep equilibrium (i.e, DEQ-full-sc, DEQ-fast-sc) models for the HSI denoising problem. The deep unrolled models were trained for $K=10$ iterations. The deep unrolling approaches are optimized only for a fixed number of iterations/layers during training, and hence increasing the number of iterations during inference deteriorates the reconstruction accuracy. On the other, the proposed deep equilibrium approaches are able to maintain or improve their performance for a wide range of iteration. Additionally, it can be seen that the proposed models notably outperform the plug-and-play approaches.
} 
  \label{fig:iteration_psnr}
\end{figure}

Focusing on the proposed methods, the Deep Equilibrium-based models, namely the DEQ-full-sc and DEQ-fast-sc consistently outperform the Deep Unrolling approaches (i.e., the DU-full-sc and the DU-fast-sc) in all cases. Furthermore, Figure \ref{fig:iteration_psnr} illustrates another great merit of the proposed Deep Equilibrium-based approaches. In more detail, the Deep Unrolling methods, both the full and fast versions are optimized for a fixed number of iterations/layers during training, and hence increasing the
number of iterations during inference deteriorates the reconstruction accuracy. On the other hand, the proposed deep equilibrium methodologies are able to maintain or improve their performance for a wide range of iterations, providing a balance between the desired computational complexity and accuracy. Note that similar results were obtained by
considering the other types of noise, however due to space
limitations, we omit the respective figures.
Additionally, it should be highlighted  that an additional difference between the deep equilibrium and deep unrolling models is the fact that the deep unrolling models provide the flexibility to employ different denoisers at each layer (i.e., non-tied weights approach) compared to the corresponding deep equilibrium architectures that are restricted to use the same denoiser at each layer (iteration). However, in our experiments (see, Figure \ref{fig:num_of_param}), we observed that the non-tied-weights deep unrolling models provided no performance gains as compared to the tied-weights deep unrolling models. The above remark is in-line with recent works e.g., \cite{rebeca}.

Among the Deep Equilibrium methods, the fast version (i.e., DEQ-fast-sc) provides slightly better reconstruction results compared to the full version, namely the DEQ-full-sc. This can be explained by considering that the examined noisy images consists of blocks with underlying strong dependencies, thus the assumption made in Section \ref{method B} that the signals in each block can be described by the same support set based on their corresponding average/centroid signal is well valid. In other words, the average/centroid signals of the blocks are, in essence, a denoised average vector that represents all noisy signals in the blocks, thus enabling the fast method to estimate more accurate supports. However, there is no an obvious winner between the Deep Equilibrium methods, thus indicating that the optimal choice in a particular application  may be problem- or setting-dependent.

\begin{table}
\centering
  \caption{Proposed methods versus Plug-and-Play (PnP) approaches.}

  \resizebox{\linewidth}{!}{
  \begin{tabular}{ccccccccl}
    \toprule
    Noise& Metrics  & PnP-full       & PnP-fast        &DU-full-sc          & DEQ-full-sc &DU-fast-sc & DEQ-fast-sc\\
    
    \midrule
     i.i.d.&PSNR  & 39.15   & 39.65   & 42.60  & 42.80             & 42.75              & \textbf{42.87} \\
     sigma 30 &SSIM  & 0.941  & 0.950    & 0.972  & 0.973             & 0.974               & \textbf{0.975} \\
         &SAM   & 0.070   & 0.067   & 0.045  & 0.042             & 0.042              & \textbf{0.041}\\
    \midrule
    i.i.d.  &PSNR  &37.73   & 37.84    & 40.22  & 40.75             & 40.62              & \textbf{40.77} \\
    sigma  50 &SSIM  &0.927   & 0.932    & 0.956  & 0.962             & 0.961              & \textbf{0.963} \\
         &SAM   &0.079   & 0.072    & 0.054  & \textbf{0.047}    & 0.050              & 0.048\\
    \midrule
    i.i.d. &PSNR  &36.21     & 36.43    & 39.14  & \textbf{39.27}   & 39.01             & 39.25 \\
    sigma  70 &SSIM  &0.901     & 0.915    & 0.947  & 0.949            & 0.948             & \textbf{0.950} \\
         &SAM   &0.089     & 0.081    & 0.056  & \textbf{0.054}   & 0.056             & 0.055\\
    \midrule
    \midrule
         &PSNR  &39.04     & 39.35         & 42.31  & 42.75     & 42.63      & \textbf{42.91} \\
non i.i.d. &SSIM   &0.948     & 0.958      &0.969  & 0.974      & 0.970      & \textbf{0.976} \\
         &SAM   &0.078     & 0.070        &0.053  & 0.051      & 0.051      & \textbf{0.049}\\
    \midrule
         &PSNR  &39.17     & 39.54       &42.15   & 42.48           & 42.30           &  \textbf{42.55}\\
non i.i.d.  &SSIM  &0.948     & 0.952    &0.965   & 0.975           & 0.970            &  \textbf{0.977}\\
+ stripes   &SAM   &0.077     & 0.065    & 0.058   & 0.052           & 0.055            & \textbf{0.052}\\
    \midrule
         &PSNR  &38.88     & 39.05      &42.18   & 42.35           & 42.25        & \textbf{42.44} \\
non i.i.d.  &SSIM  &0.940     & 0.948   &0.961   & 0.973             & 0.0971        &  \textbf{0.975}\\
+ deadline &SAM   &0.085     & 0.080    &0.060 & 0.055      & 0.058         & \textbf{0.055}\\
    \bottomrule
  \end{tabular}}
    \label{tab:results_pnp}
\end{table}

\subsection{Ablation Analysis} 
In this section, an ablation study was conducted in order to investigate
the sensitivity of the proposed methods to the most significant parameters that affect their performance.

\subsubsection{Impact of block size}

Figure \ref{fig:patch_noise} illustrates the impact of the block size on the performance of the proposed models. In particular, we note that all the proposed models are able to maintain better or competitive performance against the  state-of-the-art denoising model i.e., QRU3D \cite{QRU3D}, for a wide range of block size values. In general, larger blocks provide better performance, since the learnable regularizer (CNN network) is able to capture more accurately the dependencies of the input images.  In the absence of ground truth images, and based on the findings of Figure \ref{fig:patch_noise}, a better strategy is to use a larger value for the block size parameter. In any case, since the block size parameter is related to the spatial extent of the dependencies in the considered images, one can choose this parameter by utilizing any such prior knowledge about the input data.

\subsubsection{Impact of the Dictionary}

\begin{figure}
\centering
 \includegraphics[scale=0.45]{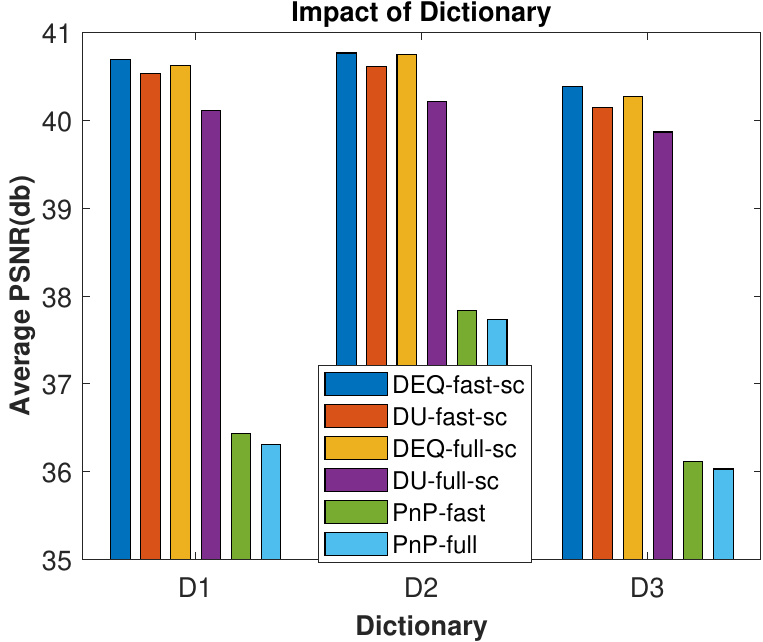}
  \caption{The impact of dictionary on the restoration performance of the proposed models under i.i.d. Gaussian noise with level $\sigma=50$, employing three different dictionaries i.e., D1, D2, D3.
}
  \label{fig:impact_of_dictionary}
\end{figure}

To explore the impact of the dictionary on the performance of the proposed deep unrolling and equilibrium methods, we conducted experiments considering the following three different dictionaries:
\begin{itemize}
    \item D1: Based on \cite{Elad2006}, the dictionary is built as an overcomplete separable version of the DCT dictionary
by sampling the cosine wave in different frequencies.
    \item D2: This dictionary is computed by following a dictionary learning approach using the noisy images in the training dataset, as detailed in Section \ref{sec_dict}, and is the option used in all previous experiments.
    \item D3: This dictionary has been derived using noisy simulated images from a different dataset i.e., the Harvard \cite{Harvard} dataset  consisting of images with $31$ spectral bands in the visible spectrum 400-700 nm. 
\end{itemize}
According to Figure \ref{fig:impact_of_dictionary}, we can deduce that the performance of the proposed models  remains competitive in all cases, regardless of the dictionary employed. Interestingly, using a generic dictionary such as the DCT  or a dictionary derived form a different dataset  does not affect significantly the performance of our approaches. In more detail, in all cases the proposed deep equilibrium models are able to exhibit better performance against the corresponding deep unrolling approaches.

The ability of the proposed models to retain competitive performance for different dictionaries can be attributed to the fact that the proposed approaches, given a fixed dictionary, are optimized end-to-end, thus all their learnable parameters are adapted to the structure of the selected dictionary. The above explanation can be confirmed by observing in Figure \ref{fig:impact_of_dictionary} that in the case of the plug-and-play approaches their performance significantly deteriorates, when the dictionary is trained from a different dataset (D3 dictionary) or it is derived from the DCT transformation (D1 dictionary), since the neural network is trained independently from the considered problem at hand and the dictionary.
Note that similar results were obtained by considering the other types of noise, however due to space limitations, we omit the respective figures.

\subsection{Discussion and Future Work} 
Apart from the superior performance of the proposed models, their major difference/advantage over the deep learning approaches e.g.,QRU3D \cite{QRU3D}, MemNet \cite{MemNet} and HSID-CNN \cite{HSID-CNN} is more fundamental. In more detail, the above mentioned techniques are designed to  tackle a specific task that is the HSI denoising problem. 
On the other hand, it must be stressed that the proposed high performance denoiser is actually a useful and interesting by-product of a more general technique.  Indeed, the primary goal of this work is to present a novel bridge between the sparse representation theory and the deep-learning models, providing an efficient methodology which preserves the generic modeling capacity
of the classical sparse representation algorithms.  This methodology could be applied to several problems which involve locally dependent signals as those appearing in hyperspectral imaging (e.g, unmixing, spatial and
spectral super-resolution, deconvolution). 

Moreover, a strong advantage of the proposed models derives form the fact that  the proposed approaches can be extended to consider learnable regularizers other than the simple CNN model in relation (\ref{eq:update3}). In particular, any state-of-the-art HSI denoiser, such as QRU3D \cite{QRU3D}, can be employed. The resulting scheme is expected to exhibit good performance in a wide range of applications, other than denoising, such as in HSI image unmixing or in super-resolution approaches. For instance, in the unmixing problem various optimization-based sparse coding algorithms \cite{sunsanltv},  provide satisfactory results. Based on the proposed optimization problem in (\ref{eq:main_problem}), i.e., 
\begin{equation}
   \underset{\mat{G}}{\arg\min}\,\,\, \frac{1}{2} \norm{\mat{Y} - \mat{D}\mat{G}}_F^2 + \mu\norm{\mat{G}}_{1,1} + \lambda \mathcal{R}(\mat{D}\mat{G})\ , 
\end{equation}
the dictionary can be replaced by a known library of endmembers and the variable $\mat{G}$ is the desired matrix of the abundances. Thus, we can derive highly interpretable deep equilibrium and unrolling networks to tackle this challenging problem. As mentioned above the unknown regularizer $\mathcal{R(\cdot)}$ can be replaced by state-of-the-art hyperspectral denoisers.

Another appealing direction, is to incorporate the dictionary
matrix into the proposed sparse coding approaches and treat
it as a learnable parameter, thus providing novel deep unrolling/equilibrium dictionary learning models. However,
this procedure entails some challenges and is left for future
work.

\begin{table}
  \caption{Average runtime (in seconds) of the proposed methods to reconstruct a HSI of size $1300 \times 1300 \times 31$  .}
    \begin{center}
    \resizebox{\linewidth}{!}{
    \begin{tabular}{ccccc}
    \toprule
         Method & DU-full-sc  & DEQ-full-sc& DU-fast-sc 2 & DEQ-fast-sc\\
        & Section \ref{DU_A} & Section \ref{DEQ_A} &  Section \ref{DU_B} &  Section \ref{DEQ_B} \\

        \midrule
        time[sec] & 13.12 & 25.03 & 4.70 & 7.45  \\
        \bottomrule
    \end{tabular} }
    \end{center}
    \label{tab:runtime}
    
\end{table}

\section{Conclusions}\label{conclusions}
In this work, a strong bridge between the sparse representation modeling and deep learning tools based on the deep equilibrium and unrolling methodologies.
 The problem of computing a sparse representation for multidimensional datasets of locally dependent signals was considered.  A regularized optimization approach was proposed, where the considered dependencies are captured using regularization terms which was properly learnt from the data. Deep equilibrium and deep unrolling based algorithms were developed for the considered problem. Extensive simulation results, in the context of hyperspectral image denoising, were provided, that demonstrated some very promising results in comparison to  plug-and-play methodologies and several recent state-of-the-art 
 denoising models.

\bibliographystyle{IEEEtran}
\bibliography{mybibfile}

\end{document}